\PassOptionsToPackage{table}{xcolor}
\documentclass{article} 
\usepackage{iclr2027_conference,times}

\usepackage{amsmath,amsfonts,bm}

\def\eqref#1{equation~\ref{#1}}

\def\1{\bm{1}}

\DeclareMathAlphabet{\mathsfit}{\encodingdefault}{\sfdefault}{m}{sl}
\SetMathAlphabet{\mathsfit}{bold}{\encodingdefault}{\sfdefault}{bx}{n}

\usepackage{hyperref}
\usepackage{url}
\usepackage{algorithm}
\usepackage{algorithmic}
\usepackage{amsthm}
\usepackage{booktabs}
\usepackage{adjustbox}
\usepackage{amsmath}
\usepackage{amssymb}    
\usepackage{colortbl}
\usepackage{bm}
\usepackage{mathtools}
\usepackage{enumitem}
\usepackage{wrapfig}
\usepackage{arydshln}
\usepackage{multirow}
\usepackage{caption} 
\usepackage[table]{xcolor}
\usepackage{graphicx}
\usepackage{tabularx}
\usepackage{array}

\newcolumntype{a}{>{\columncolor{gg}}c}

\theoremstyle{definition}

\theoremstyle{remark}

\definecolor{figgreen}{RGB}{59, 125, 35}
\definecolor{figorange}{RGB}{215, 137, 0}
\definecolor{figred}{RGB}{192, 0, 0}
\definecolor{figblue}{RGB}{48, 132, 194}
\definecolor{mygray}{gray}{0.6}
\definecolor{Red}{rgb}{0.8,0,0}
\definecolor{Blue}{rgb}{0,0,0.8}
\definecolor{gg}{gray}{0.92}
\definecolor{tabred}{RGB}{230, 10, 10}
\definecolor{tabblue}{RGB}{10, 10, 230}
\definecolor{citeblue}{rgb}{0.15,0.35,0.55}
\definecolor{linkred}{rgb}{0.9,0,0}
\definecolor{Blue9}{rgb}{0.1,0.3,0.95}
\hypersetup{
    colorlinks = true,
    citecolor = citeblue,
    linkcolor = linkred,
    urlcolor = citeblue
}

\usepackage{tcolorbox}
\tcbuselibrary{skins} 
\tcbuselibrary{listings}
\tcbuselibrary{breakable}

\usepackage[T1,OT1]{fontenc}
\ifXeTeX
    \IfFileExists{lmmono8-regular.otf}{}{}
    ="[lmmono8-regular.otf]" at 7pt
    \newcommand{\agentpromptfont}{\fontsize{7}{8.5}\selectfont\agentprompttypewriter}
\else
    \newcommand{\agentpromptfont}{\fontencoding{T1}\fontfamily{cmtt}\fontsize{7}{8.5}\selectfont}
\fi

\ifXeTeX
    ="[lmmono8-regular.otf]" at 9pt
    \newcommand{\contactfont}{\fontsize{9}{10}\selectfont\contacttypewriter}
\else
    \newcommand{\contactfont}{\fontencoding{T1}\fontfamily{cmtt}\fontsize{9}{10}\selectfont}
\fi

\newtcolorbox{promptbox}{
    colback=gray!5,
    colframe=gray!70,
    boxrule=0.8pt,
    arc=2mm,
    left=10pt,
    right=10pt,
    top=10pt,
    bottom=10pt,
    width=\linewidth,
    fontupper=\agentpromptfont,
    enhanced,
    breakable,
    lines before break=4,
    before skip=8pt,
    after skip=0pt,
}
\lstdefinestyle{agentprompttext}{
    language={},
    basicstyle=\agentpromptfont,
    columns=fullflexible,
    keepspaces=true,
    showstringspaces=false,
    upquote=true,
    breaklines=true,
    breakatwhitespace=true,
    breakautoindent=false,
    breakindent=0pt,
    aboveskip=0pt,
    belowskip=0pt,
    tabsize=2,
    literate={—}{{\textemdash}}1 {–}{{\textendash}}1 {…}{{\ldots}}1
             {’}{{\textquoteright}}1 {‘}{{\textquoteleft}}1,
}

\title{AutoSciBench: Autonomous Benchmark Generation for Evaluating Scientific Agents}

\author{Dongki Kim$^{1\,2}$\thanks{Work done during internship at Genentech.}$\enspace$, Namkyeong Lee$^{1}$\thanks{Corresponding Authors.}$\enspace$, Surag Nair$^1$, Carl Edwards$^1$, Xiner Li$^1$, \\\textbf{Edward De Brouwer}$^1$, \textbf{Jenna Lynn Collier}$^1$, \textbf{Sung Ju Hwang}$^2$, \textbf{Gabriele Scalia}$^1$,  \\\textbf{Ehsan Hajiramezanali}$^{1\,\dagger}$ \\[0.5em]
$^{1}$ Genentech \quad $^{2}$ KAIST \\[0.25em]
{\contactfont cleverki@kaist.ac.kr} \quad {\contactfont \{lee.nam\_kyeong, hajiramezanali.ehsan\}@gene.com}\\[0.25em]}

\iclrfinalcopy 
\begin{document}

\maketitle
\lhead{Preprint}

\begin{abstract}
As agents rapidly evolve, existing benchmarks can become saturated, limiting their ability to distinguish capabilities and reveal remaining failure modes. Particularly in scientific domains, constructing and updating benchmarks requires substantial time, labor, and domain expertise, making it difficult to keep evaluation aligned with advances in agent capabilities.  We address this challenge by investigating whether scientific-agent benchmarks can be automatically generated and iteratively adapted as agent capabilities evolve. We introduce AutoSciBench, a framework that represents each task as a high-level concept specifying the scientific domain, data modality, and required reasoning approach, together with a low-level recipe specifying how the question, environment, and ground-truth answer are constructed and verified. Agents attempt to solve each task, producing solver trajectories and corresponding judge feedback which AutoSciBench uses to revise the recipe or concept, closing observed shortcuts and shifting tasks toward raw-data re-examination, interpretation of intermediate results, and evidence integration. Experience distilled from completed refinement trajectories further guides new concept generation, allowing lessons from earlier task refinement to inform subsequent benchmark construction. Starting from existing benchmarks, we evaluate AutoSciBench across computational biology, materials science, and clinical imaging. Generated benchmarks reduce average solver accuracy by 22.4 and 25.5 percentage points relative to the human-curated benchmarks in computational biology and materials science, respectively, while generated tasks receive higher average quality ratings across all three domains. Further analyses show that iterative refinement increases task difficulty and prior refinement experience guides more challenging task generation. Together, these results suggest that scientific-agent evaluation can adapt as agent capabilities advance.
\end{abstract}
\section{Introduction}

Benchmarking has played a critical role in evaluating agents, providing a controlled basis for characterizing their capabilities and failure modes. As agent capabilities have advanced rapidly~\citep{anthropic2026opus47,anthropic2026opus48,anthropic2026opus5,openai2026gpt54,openai2026gpt55,openai2026gpt56,openai2026gpt6}, evaluation has struggled to keep pace. This challenge is particularly pronounced for scientific agents: constructing and updating  benchmarks requires substantial time, labor, and domain expertise, while existing benchmarks can quickly become saturated. This growing gap between agent capabilities and benchmark construction is particularly consequential in scientific domains such as drug discovery, where reliable agent analyses are critical for research. 

A natural approach to addressing this challenge is to automate the benchmark construction process itself. Recent work has explored this direction, primarily focusing on generating questions for LLM evaluation~\citep{butt2024benchagents,li2025autobencher,wang2025benchmarkselfevolving,yuan2026benchmaker}. Extending this approach to scientific agents requires constructing not only questions, but also the scientific data and evaluation settings needed to answer them. The question and data should jointly assess how agents acquire information and reason from scientific evidence. Thus, the challenge lies in translating desired scientific capabilities into concrete, verifiable evaluation tasks.
This raises a broader question: can scientific-agent evaluations be automatically constructed and improved in response to the capabilities and failure modes of the agents they evaluate?

\begin{figure}[t]
    \centering
    \includegraphics[width=\linewidth]{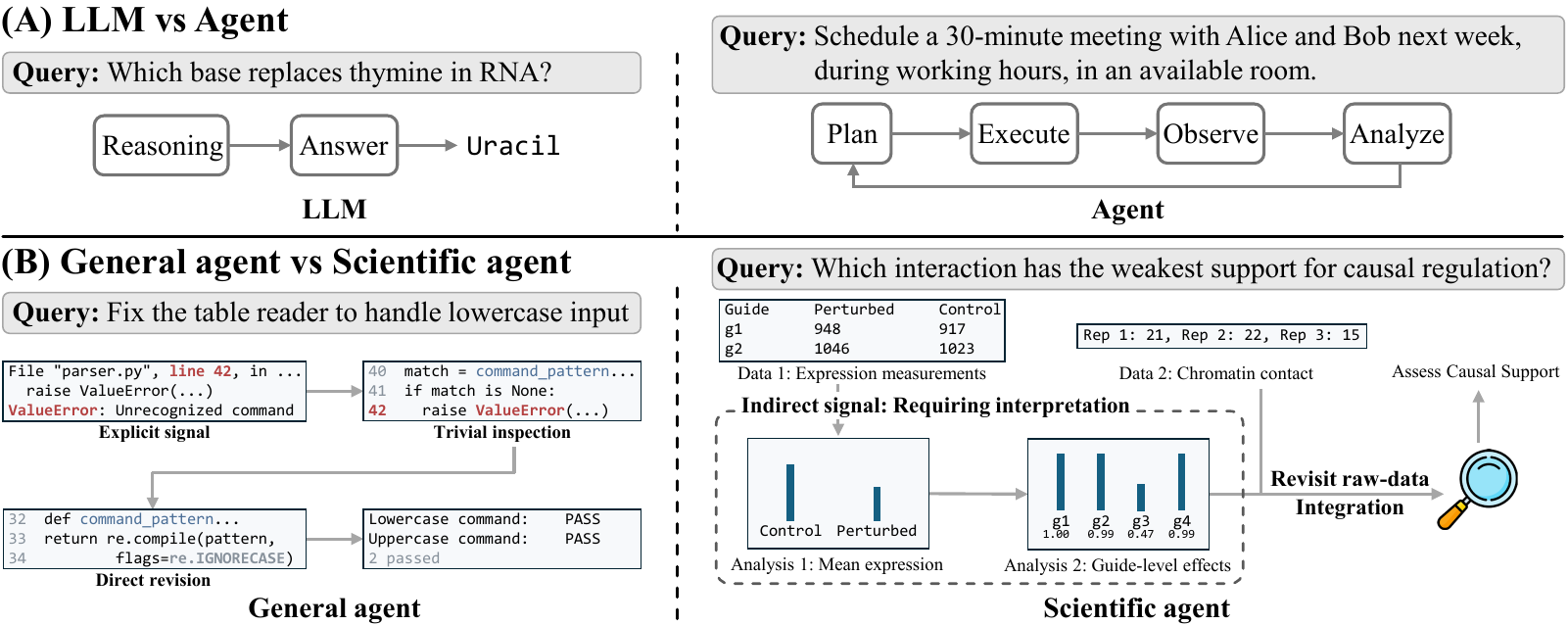}
    \vspace{-0.25in}
    \caption{Conceptual illustration of two key distinctions: LLM vs Agent and General agent vs Scientific agent. (A) An LLM answers using the provided context and parametric knowledge, whereas an agent acquires additional information through interaction with external data, tools, and environments, involving planning, tool execution, observation, and analysis. (B) In general agent tasks, execution errors and observable state changes often provide explicit feedback for revising a solution. In scientific analysis tasks, intermediate outputs typically provide evidence about a quantity or hypothesis of interest rather than explicit confirmation that the analysis is correct, requiring agents to re-examine raw data, validate intermediate results, and integrate evidence across analysis steps.}
    \label{fig:conceptual_comparison}
    \vspace{-0.25in}
\end{figure}

To identify the capabilities that a scientific-agent evaluation should assess, we highlight two key distinctions: LLMs vs Agents and General agent vs Scientific agent. \textbf{First, Information Acquisition through Interaction} (Fig.~\ref{fig:conceptual_comparison}(A): LLM vs Agent). In conventional LLM tasks, the information needed to answer a question is available in the input context or the model's parametric knowledge. In contrast, agentic tasks require acquiring additional information through interaction with the environment (including data and tools), which involves planning, tool execution, observation, and analysis. Therefore, constructing scientific evaluations requires specifying both the environment and the operations needed to acquire answer-relevant information. 

\textbf{Second, Scientific Reasoning from Indirect Evidence} (Fig.~\ref{fig:conceptual_comparison}(B): General agent vs Scientific agent). Beyond acquiring information, scientific agents should determine what conclusions are supported by the resulting observations. In code and web environments, agents often receive explicit feedback through execution errors or observable state changes, which helps them diagnose failures and revise their plans~\citep{jimenez2024swebench,xie2024osworld,zhou2024webarena}. In scientific analysis, however, intermediate outputs typically provide evidence about a quantity or hypothesis of interest, rather than explicit feedback on whether the analysis is correct. For example, an agent should analyze the raw expression measurements to examine guide-level effects beyond mean expression.
These results should then be interpreted with chromatin-contact evidence to assess support for causal regulation. Scientific-agent tasks should therefore assess whether agents can re-examine raw data, validate intermediate results, and integrate evidence across multiple analysis steps.

Motivated by these requirements, we investigate the autonomous generation and refinement of scientific-agent benchmarks by developing AutoSciBench. We organize task construction around two levels of planning.
The high-level plan, termed a \textit{concept}, specifies the scientific domain, data modality, and analytical operations to be evaluated. The low-level plan, termed a \textit{recipe}, translates these requirements into a concrete question and data design, specifying which information is provided directly and which must be recovered through interaction.
We then implement these plans to produce a question, associated scientific data files, and a ground-truth answer. However, a well-formed concept and recipe do not guarantee that the resulting task elicits the intended scientific reasoning: an agent may still obtain the correct answer through a shortcut, such as a single tool output. To better assess scientific reasoning from indirect evidence, we introduce feedback-guided iterative refinement, which uses solver behavior to revise the recipe or the underlying concept, making the resulting task progressively more dependent on raw-data re-examination, interpretation of intermediate results, and evidence integration. Finally, we summarize the refinement histories and generate subsequent concepts based on these histories, allowing the construction of more challenging tasks. 

Using AutoSciBench, we investigate automatic benchmark construction across three scientific fields: computational biology, materials science, and clinical imaging. Our results show that the generated benchmarks are more challenging than the original manually-curated benchmarks, while also receiving higher average quality scores across all three domains. Further analyses demonstrate the effectiveness of the key components: iterative refinement progressively increases task difficulty by introducing requirements for data re-examination and interpretation, while experience-guided concept generation supports the construction of increasingly challenging tasks.
Together, these findings show that scientific-agent benchmarks can be automatically constructed and iteratively adapted to rapidly evolving agent capabilities.

\section{Related Work}
\vspace{-0.05in}

\paragraph{Human-curated Benchmarks for LLMs and Agents}
Numerous LLM benchmarks evaluate a broad range of capabilities using fixed questions
that can be answered from the provided input or the model's parametric knowledge, including general knowledge, mathematical reasoning, code generation, and scientific reasoning~\citep{hendrycks2021mmlu,cobbe2021gsm8k,chen2021codex,rein2024gpqa}. 
As LLMs have evolved into agents, evaluation has shifted toward executable tasks requiring planning, tool use, and interaction with external environments.
SWE-bench evaluates repository-level software engineering, while WebArena and OSWorld evaluate agents in web and computer environments, respectively~\citep{jimenez2024swebench,zhou2024webarena,xie2024osworld}. LAB-Bench, ScienceAgentBench, and CompBioBench further evaluate scientific reasoning, data analysis, and domain-specific tool use~\citep{laurent2024labbench,chen2025scienceagentbench,nair2026compbiobench}. 
However, constructing these benchmarks requires substantial human effort and domain expertise, particularly in scientific domains where reliable evaluation often depends on costly expert annotation \citep{chan2026dcws}, making it difficult to continuously update evaluations as agent capabilities advance.
\paragraph{Automatic Benchmark Generation}
Existing automatic benchmark generation uses LLMs to reduce the human effort involved in constructing evaluation tasks. Specifically, AutoBencher and BenchMaker generate evaluation instances from declarative objectives or predefined quality criteria~\citep{li2025autobencher,yuan2026benchmaker}. On the other hand, BenchAgents and Benchmark Self-Evolving organize benchmark construction into multiple stages or expand existing benchmarks through predefined transformations~\citep{butt2024benchagents,wang2025benchmarkselfevolving}. For safety evaluation, S-Eval generates and selects prompts based on a hierarchical risk taxonomy~\citep{yuan2024seval}. However, these methods primarily focus on LLM evaluation rather than scientific-agent tasks that require information acquisition through interaction and scientific reasoning from indirect evidence. 
Our framework addresses these requirements through high-level and low-level planning for question and data construction, together with feedback-guided refinement based on solver behavior.
\section{AutoSciBench}
\label{sec: method}

\begin{figure}[t]
    \centering
    \includegraphics[width=\linewidth]{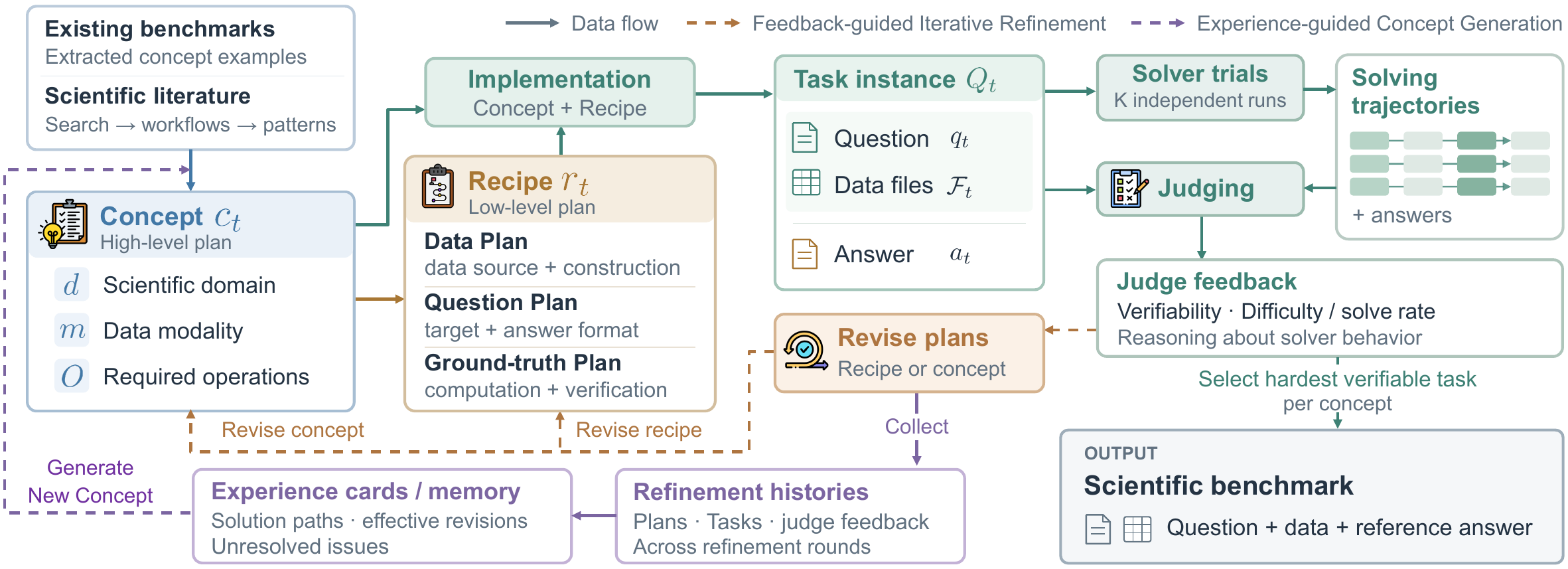}
    \vspace{-0.25in}
    \caption{Overview of AutoSciBench. Existing benchmarks and scientific literature inform concept generation, followed by recipe planning and task implementation. Solvers attempt the generated tasks, while the judge provides feedback on the generated tasks. The resulting feedback guides revisions to the recipe or concept, producing tasks that are more verifiable and challenging. Refinement histories are summarized into experience memory to guide subsequent concept generation.}
    \label{fig:method_overview}
    \vspace{-0.05in}
\end{figure}

In this section, we introduce AutoSciBench, a framework for automatically constructing and refining analytical benchmarks for scientific-agent evaluation (Fig.~\ref{fig:method_overview}).
Given a user-specified scientific field, AutoSciBench operationalizes benchmark generation as a repeated cycle of concept generation, recipe planning, task implementation, solver-based evaluation, and refinement. Solver feedback guides revisions to the concept or recipe, while experience distilled from completed refinement trajectories informs the generation of subsequent concepts.

\subsection{Concept: High-level plan for evaluation requirements}

To represent the evaluation requirements of a scientific-agent task, we define a \textit{concept} as a high-level plan that specifies the types of scientific data and analytical capabilities to be evaluated. Specifically, we represent each concept as a tuple $c=(d,m,O)$ with three dimensions: the scientific domain $d$, the data modality $m$, and the set of required operations $O$. The domain $d$ specifies the scientific focus of an individual task within the selected field, while the data modality $m$ specifies the type of data the agent should analyze. Within this domain and modality, the required operations specify what the agent is expected to infer, estimate, compare, or integrate from the data. 
This abstraction separates what should be evaluated from how it is concretely instantiated, allowing multiple tasks with different questions and data to target the same analytical capability.

\paragraph{Initial Concept Generation}
For a specified field, AutoSciBench derives source priors from two complementary external sources to generate concepts: existing scientific-agent benchmarks and scientific literature. For each seed question and its associated data files, an agent extracts the scientific domain, data modality, and required operations to form a concept. The resulting benchmark-derived concepts provide examples of existing evaluation targets.

To capture analytical procedures beyond those represented in existing benchmarks, AutoSciBench searches for relevant publications within the specified field and extracts workflows describing how scientific objectives are addressed through a sequence of analytical steps. Then, the extracted workflows are generalized into reusable patterns that preserve recurring  analytical steps and decision points while abstracting away study-specific details. The resulting workflows provide concrete analytical examples, while the patterns capture common structures across these examples.

The extracted concepts, workflows, and patterns jointly serve as source-grounded references for new concept generation.
These sources provide examples of evaluation requirements and scientific procedures that inform the domain, data modality, and required operations of each new concept. The resulting concepts specify the analytical requirements to be evaluated, while leaving data selection and input-file construction to recipe planning (Fig.~\ref{fig:generation_example}(A)).

\subsection{Recipe: Low-level plan for Task construction}

A concept specifies what analytical capability should be evaluated, but leaves open how it should be instantiated in a concrete task. We define a \textit{recipe} $r$ as a low-level plan that translates these requirements into a specific task designed to assess information acquisition through interaction.

\paragraph{Recipe Planning}
Given a concept $c$, an agent develops a recipe $r$ that links the required operations to the information the solver is expected to obtain from the scientific data. The agent plans the question and associated data together, determining which information is stated in the question and which should be acquired through the intended analysis. The resulting recipe is designed so that answering the question depends on observations obtained through the intended analytical operations, rather than solely on the question text or the model's parametric knowledge (Fig.~\ref{fig:generation_example}(B)).

\begin{figure}[t]
    \centering
    \begingroup
    \color{black}
    \definecolor{exampledomain}{RGB}{0,90,200}
    \definecolor{examplemodality}{RGB}{200,40,40}
    \definecolor{exampleoperation}{RGB}{0,125,65}
    \newcommand{\exdomain}[1]{\textcolor{exampledomain}{\textbf{#1}}}
    \newcommand{\exmodality}[1]{\textcolor{examplemodality}{\textbf{#1}}}
    \newcommand{\exoperation}[1]{\textcolor{exampleoperation}{\textbf{#1}}}
    \newcommand{\exfield}[1]{\par\smallskip\textbf{#1}\par\nobreak}
    \fontsize{8.5}{10.5}\selectfont
    \setlength{\tabcolsep}{0pt}
    \renewcommand{\arraystretch}{1.05}
    \begin{tabular}{@{}p{0.28\linewidth}@{\hspace{0.02\linewidth}}p{0.33\linewidth}@{\hspace{0.02\linewidth}}p{0.35\linewidth}@{}}
        \toprule
        \textbf{(A) Concept}: \textit{High-level plan} &
        \textbf{(B) Recipe}: \textit{Low-level plan} &
        \textbf{(C) Task}: \textit{QA + Data} \\
        \midrule
        \begin{minipage}[t]{\linewidth}
            \raggedright
            \textbf{Scientific domain}\par
            \exdomain{Crystal chemistry / perovskite structure prediction}

            \exfield{Data modality}
            \exmodality{Ionic radii} at specified coordination numbers

            \exfield{Required operations}
            \begin{itemize}[leftmargin=9pt,labelsep=3pt,nosep]
                \item \exoperation{Compute tolerance factor}
                \item \exoperation{Compute octahedral factor}
                \item \exoperation{Classify the structure}
            \end{itemize}
        \end{minipage}
        &
        \begin{minipage}[t]{\linewidth}
            \raggedright
            \textbf{Data plan}\par
            Select an \exdomain{ABX\textsubscript{3} compound}.
            Retrieve \exmodality{ionic radii} and encode classification rules in two CSV files.

            \exfield{Question plan}
            Ask for the \exoperation{structural class}.
            Do not provide formulas or computed values.

            \exfield{Ground-truth plan}
            \exoperation{Compute both factors}, \exoperation{apply the rules}, and verify the answer.
        \end{minipage}
        &
        \begin{minipage}[t]{\linewidth}
            \raggedright
            \textbf{Question}\par
            Using the \exmodality{ionic radii} in \texttt{ions.csv} and the rules in
            \texttt{classification} \texttt{\_criteria.csv}, \exoperation{predict the structural class}
            of \exdomain{CsPbBr\textsubscript{3}}.

            \exfield{Data: \texttt{ions.csv}}
            {
            \fontsize{7.5}{8.5}\selectfont
            \renewcommand{\arraystretch}{0.82}
            
            \begin{tabular}{@{}l@{\hspace{15pt}}c@{\hspace{15pt}}c@{\hspace{15pt}}r@{}}
                \toprule
                \texttt{Site} &
                \texttt{Ion} &
                \texttt{CN} &
                \exmodality{Radius (\r{A})} \\
                \midrule
                \texttt{A} & \texttt{Cs} & \texttt{12} & \texttt{1.88} \\
                \texttt{B} & \texttt{Pb} & \texttt{6}  & \texttt{1.19} \\
                \bottomrule
            \end{tabular}
            }

            \par\smallskip
            \textbf{Answer:} distorted \exdomain{perovskite}
        \end{minipage}
        \\
        \bottomrule
    \end{tabular}
    \endgroup
    \vspace{-0.1in}
    \caption{An abbreviated example of task construction from concept definition to recipe planning and task implementation. Matching colors indicate how the scientific domain, data modality, and required operations are reflected in the recipe and the resulting question, answer, and data files.}
    \label{fig:generation_example}
\end{figure}

\subsection{Task Implementation and Iterative Refinement}

We first implement the concept and recipe to construct an initial task. However, the resulting task may be solvable through a single tool output without requiring the intended scientific reasoning, such as re-examining raw data or integrating intermediate results. We therefore introduce feedback-guided iterative refinement to identify such solution paths and revise the concept and recipe based on observed solver behavior.

\paragraph{Task Implementation}
For a single refinement run, we denote the current concept and recipe at round $t$ by $c_t$ and $r_t$, respectively, with $t=0$ corresponding to the generated concept and its initial recipe. At round $t$, an agent implements the recipe $r_t$ for concept $c_t$ to produce a task instance $Q_t=(q_t,\mathcal{F}_t,a_t)$, where $q_t$ is the question, $\mathcal{F}_t$ is the set of associated scientific data files, and $a_t$ is the ground-truth answer. Specifically, the agent obtains or constructs the data, computes and verifies the answer, and writes the question according to the recipe. The question identifies the analysis target, input files, and required answer format without providing a solution procedure. The resulting task instantiates the concept and recipe as a concrete evaluation instance, comprising a question, associated data files, and a ground-truth answer (Fig.~\ref{fig:generation_example}(C)).

\paragraph{Solving and Judging}
For each implemented task $Q_t$, we run $K$ independent solver trials and collect their trajectories and final answers. Based on the solver trajectories and final answers, the judge assesses two aspects: (i) verifiability, whether the question and data support a well-defined, gradable answer, and (ii) difficulty, how reliably the solver derives the correct answer across the $K$ trials, measured by the empirical solve rate.
The solver trajectories also help identify solution paths that bypass the intended validation or evidence integration, as well as analytical errors that lead to incorrect answers. 
The resulting judge verdict contains the verifiability and difficulty ratings, the solve count, and textual reasoning, which together provide feedback for subsequent refinement.

\paragraph{Feedback-guided Refinement}
If $Q_t$ is not both verifiable and hard, the agent uses judge feedback, together with the construction records, to revise its recipe or underlying concept.
Recipe revision addresses shortcomings in how the current evaluation target is instantiated, while concept revision changes the target itself when its analytical requirements fail to elicit the intended scientific reasoning.
Specifically, recipe revision preserves the concept, i.e. $c_{t+1}=c_t$, and produces a revised recipe $r_{t+1}$ that changes the data and question design, 
so that solving the task requires additional validation, interpretation, or evidence integration beyond a direct intermediate output.
Concept revision instead updates $c_t$ to $c_{t+1}$ and develops a new recipe $r_{t+1}$ for the revised analytical requirements, allowing subsequent tasks to require other forms of reasoning and operations 
that the previous concept did not require. The feedback also guides corrections to ambiguity and implementation problems that prevent reliable evaluation. After either type of revision, the recipe $r_{t+1}$ is implemented to construct the next task instance $Q_{t+1}$. The resulting task undergoes the same solving and judging process. The concepts, recipes, implemented tasks, implementation-failure records, and judge verdicts accumulated across rounds form the \emph{refinement history} of a run. These histories are retained to guide subsequent concept generation, as described next.

\vspace{-0.03in}
\subsection{Experience-guided Concept Generation}
\vspace{-0.03in}
Feedback-guided refinement improves a task within a given concept, but the resulting lessons are valuable beyond that individual task. Without reusing them, concept generation may repeatedly propose evaluation targets that lead to similar shortcuts or other weaknesses encountered in earlier tasks. We therefore distill refinement histories into reusable experience that guides subsequent concept generation, allowing benchmark construction to build on both successful revisions and unresolved failure modes.
These histories provide observations about which task designs admitted direct solution paths and how revisions changed solver behavior.

\vspace{-0.03in}
\paragraph{Experience Construction}
After refinement terminates for a concept, an agent reviews the construction and evaluation records across rounds, including the task designs and judge feedback on solver behavior. It summarizes the designs that permitted direct answers, the revisions associated with changes in solver behavior, and unresolved issues affecting verifiability or difficulty. These observations are organized into an \emph{experience card} containing reusable guidance for subsequent concept generation. The accumulated experience cards form the framework's \emph{experience memory}. Experience is retained both when refinement produces a hard, verifiable task and when it does not.

\vspace{-0.03in}
\paragraph{Concept Generation with Experience}
While initial concept generation relies on external sources, subsequent generation also incorporates experience from earlier refinement histories. The agent uses the source priors together with selected cards from the experience memory to propose new concepts, drawing on observed limitations and useful revisions from previous task constructions. Previously generated concepts are also provided to discourage duplicate proposals. Each new concept then undergoes recipe planning, task implementation, and iterative refinement, and the resulting experience is retained to inform subsequent generation.

\vspace{-0.03in}
\section{Experimental Results}
\label{sec: experiments}
\vspace{-0.03in}

We evaluate AutoSciBench by assessing the difficulty and quality of the generated tasks across three different scientific fields: computational biology, materials science, and clinical imaging.
Experiments allow examining three core questions: (1) whether automatic benchmark generation can produce challenging and useful scientific-agent evaluations, (2) whether feedback-guided iterative refinement can systematically improve generated tasks, and (3) whether experience from previous refinement trajectories can improve subsequent benchmark generation.

\vspace{-0.03in}
\subsection{Benchmark Generation Details}
We leverage CompBioBench~\citep{nair2026compbiobench}, MatTools~\citep{liu2025mattools}, and MedCTA~\citep{ashraf2026medcta} as seed benchmarks for computational biology, materials science, and clinical imaging, respectively. For each domain, we infer prior concepts from 100 randomly selected seed questions and complement them with analytical workflows extracted from the corresponding scientific literature. We use Claude Code CLI with Claude Opus 4.8 and Codex CLI with GPT-5.6 Sol as generators, with each generator also serving as the solver during refinement. We construct 50 tasks per domain through five experience-guided concept-generation iterations, each producing 10 concepts. For each concept, we construct and iteratively refine a task over up to 10 rounds, including the initial construction round. To construct the final benchmark, we select one task from each refinement run, prioritizing higher difficulty ratings. Additional experimental details are provided in Appendix~\ref{app:experimental_details}.

\vspace{-0.02in}
\subsection{Model Performance on Original and Generated Benchmarks}

\vspace{-0.02in}
\paragraph{Evaluation Setting}
We provide each agent with a question and its associated data files and ask it to derive an answer by analyzing the files. We evaluate performance using accuracy and for the original CompBioBench, we report scores from the official leaderboard. For more details about configurations, please refer to Section~\ref{app:gen_infer_configuration}. The results are presented in Table~\ref{tab:main_results}.

\begin{table*}[t]
    \centering
    \caption{Model performance on the benchmarks curated by humans and generated by Claude Opus 4.8 and GPT-5.6 Sol across the three scientific domains. Rows indicate solvers, and columns indicate models used during benchmark generation. $^{*}$ denotes a leaderboard entry obtained at a reasoning effort other than high, or one whose effort is not reported. The performance on the human-curated clinical imaging benchmark is omitted due to its LLM evaluation rather than exact match.}
    \vspace{-0.1in}
    \label{tab:main_results}
    \setlength{\tabcolsep}{5pt}
    \renewcommand{\arraystretch}{1.25}
    \resizebox{\textwidth}{!}{%
    \begin{tabular}{lcccccccc>{\columncolor{black!8}}c>{\columncolor{black!8}}c}
        \toprule
        & \multicolumn{3}{c}{\textbf{Computational Biology}}
        & \multicolumn{3}{c}{\textbf{Material Science}}
        & \multicolumn{2}{c}{\textbf{Clinical Imaging}}
        & \multicolumn{2}{>{\columncolor{black!8}}c}{\textbf{Avg.}} \\
        \cmidrule(lr){2-4}\cmidrule(lr){5-7}\cmidrule(lr){8-9}\cmidrule(lr){10-11}
        & Human & Opus 4.8 & GPT-5.6 Sol
        & Human & Opus 4.8 & GPT-5.6 Sol
        & Opus 4.8 & GPT-5.6 Sol
        & Opus 4.8 & GPT-5.6 Sol \\
        \midrule
        Claude Opus 5 & 94.0          & 72.7 & 88.7 & 97.7 & 90.0 & 92.7 & 82.6 & 85.3 & 81.8 & 88.9 \\
        Claude Opus 4.8 & 90.4$^{*}$  & 50.7 & 80.3 & 96.0 & 62.0 & 88.0 & 74.2 & 78.7 & 62.3 & 82.3 \\
        Claude Opus 4.7 & 79.5$^{*}$  & 46.0 & 60.7 & 97.0 & 63.3 & 70.7 & 73.5 & 64.7 & 60.9 & 65.4 \\
        Claude Sonnet 5 & 76.0$^{*}$  & 59.3 & 81.3 & 96.7 & 83.3 & 84.5 & 68.2 & 78.0 & 70.3 & 81.3 \\
        Claude Haiku 4.5 & 34.0$^{*}$ & 26.7 & 10.7 & 93.7 & 21.3 & 29.3 & 55.7 & 44.0 & 34.6 & 28.0 \\
        \cdashline{1-11}[2pt/1pt]\rule{-2pt}{2.5ex}
        GPT-6 Astra & 98.0$^{*}$      & 81.0 & 96.0 & 96.7 & 87.3 & 96.0 & 74.2 & 90.0 & 80.8 & 94.0 \\
        GPT-5.6 Sol & 91.7            & 66.9 & 81.2 & 96.3 & 85.3 & 74.7 & 80.3 & 80.7 & 77.5 & 78.9 \\
        GPT-5.6 Terra & 64.0$^{*}$    & 54.7 & 78.5 & 94.7 & 63.3 & 74.7 & 73.5 & 74.0 & 63.8 & 75.7 \\
        GPT-5.6 Luna & 84.0           & 54.4 & 77.7 & 92.7 & 66.0 & 76.0 & 61.4 & 70.7 & 60.6 & 74.8 \\
        GPT-5.5 & 86.7                & 61.5 & 86.0 & 95.7 & 80.0 & 89.3 & 67.4 & 78.7 & 69.6 & 84.7 \\
        \cdashline{1-11}[2pt/1pt]\rowcolor{black!8}\rule{-2pt}{2.5ex}
        
        \textbf{Avg.} & 79.8 & 57.4 & 74.1 & 95.7 & 70.2 & 77.6 & 71.1 & 74.5 & 66.2 & 75.4 \\
        \bottomrule
    \end{tabular}%
    }
\end{table*}

\vspace{-0.02in}
\paragraph{Generating Challenging Tasks across Solvers}
Across computational biology and materials science, solvers generally achieve lower accuracy on the generated benchmarks than on the human-curated benchmarks, showing that AutoSciBench generates tasks that remain challenging for current agents. This tendency also holds for models other than the solver used during generation, suggesting that the generated difficulty transfers beyond the specific generator-solver pair. Notably, in materials science, solver accuracy is concentrated near the ceiling on the human-curated benchmark but varies substantially on the generated benchmarks. 
This suggests that the generated tasks can help reveal performance differences that are less apparent on the original benchmarks.

\vspace{-0.01in}
\paragraph{Generator-wise Task Difficulty}
Claude-Opus-4.8-generated benchmarks show lower average solver accuracy than GPT-5.6-Sol-generated benchmarks, while, as a solver, GPT-5.6 Sol generally outperforms Claude Opus 4.8 across benchmarks. This contrast suggests that stronger solving performance does not necessarily correspond to the ability to generate more challenging tasks.

\vspace{-0.01in}
\paragraph{Benchmark Updating for Evolving Agents}
We observe that models released after the generator models achieve substantially higher accuracy on the generated benchmarks by large margins: Opus 5.0 outperforms Opus 4.8 by 19.5pp and GPT-6 Astra outperforms GPT-5.6 Sol by 15.1pp on their respective generated benchmarks. 
These results illustrate how improvements in agent capabilities can rapidly reduce the challenge posed by a fixed benchmark, motivating continued benchmark updating. AutoSciBench supports this process by incorporating newer agents into the task-generation and refinement loop, allowing benchmarks to evolve alongside the agents they evaluate.

\vspace{-0.01in}
\subsection{Benchmark Quality Assessment}
\vspace{-0.01in}

\paragraph{Evaluation Setting}
We assess the quality of the original and generated benchmarks using two judges, Claude Opus 4.8 and GPT-5.6 Sol, each receiving the question and associated data files as inputs. We define three criteria to assess different aspects of scientific-agent tasks. \textit{Verifiability} measures whether the question specifies an unambiguous answer that can be checked reproducibly from the provided inputs. \textit{Usefulness} measures whether the question assesses a scientifically meaningful task or workflow capability relevant to real-world research. \textit{Fitness} measures suitability for agent evaluation based on the need for observation, planning, tool execution, interpretation of intermediate results, or multi-step adaptation. Each criterion is scored on a scale from 1 (very poor) to 5 (excellent). For the evaluation prompt, please refer to Fig.~\ref{fig:prompt_quality_assessment} in Appendix~\ref{app:prompt_quality_assessment}.

\vspace{-0.01in}
\paragraph{Results}
As shown in Table~\ref{tab:quality_overall}, the generated benchmarks receive higher average quality scores than the human-curated benchmarks from both judges across all three domains. GPT-5.6-Sol-generated benchmarks achieve the highest average scores, with high fitness scores across all domains, indicating that the judges identify stronger requirements for planning, tool use, and multi-step adaptation. Opus-4.8-generated tasks achieve higher usefulness scores in computational biology and materials science, indicating greater relevance to scientifically meaningful tasks and workflows in these domains. Together, these results suggest that the increased difficulty of the generated tasks does not come at the expense of the measured benchmark-quality criteria. Evaluator-specific results and inter-judge agreement are reported in Appendix~\ref{app:additional_quality}.

\vspace{-0.01in}
\subsection{Effect of Feedback-guided Iterative Refinement}
\vspace{-0.01in}

\begin{table*}[t]
    \centering
    \caption{Benchmark quality assessment on a 1-5 scale across three criteria: Verifiability (Verif.), Usefulness (Useful.), and Fitness (Fit.), higher is better. Each question is scored by two independent LLM judges, Claude Opus 4.8 and GPT-5.6 Sol, and we report the average score across both judges.}
    \vspace{-0.1in}
    \label{tab:quality_overall}
    \setlength{\tabcolsep}{6pt}
    \renewcommand{\arraystretch}{1.25}
    \resizebox{\textwidth}{!}{%
    \begin{tabular}{lcccacccaccca}
        \toprule
        & \multicolumn{4}{c}{\textbf{Computational Biology}}
        & \multicolumn{4}{c}{\textbf{Material Science}}
        & \multicolumn{4}{c}{\textbf{Clinical Imaging}} \\
        \cmidrule(l{2pt}r{2pt}){2-5}
        \cmidrule(l{2pt}r{2pt}){6-9}
        \cmidrule(l{2pt}r{2pt}){10-13}
        & Verif. & Useful. & Fit. & Avg.
        & Verif. & Useful. & Fit. & Avg.
        & Verif. & Useful. & Fit. & Avg. \\
        \midrule
        Human & 2.43 & 3.62 & 3.50 & 3.18 & \textbf{4.76} & 1.49 & 1.39 & 2.54 & 2.38 & 2.59 & 1.72 & 2.23 \\
        \cdashline{1-13}[2pt/1pt]\rule{-2pt}{2.5ex}
        Opus 4.8 & 3.08 & \textbf{4.27} & 4.42 & 3.93 & 3.49 & \textbf{4.17} & 3.96 & 3.87 & 2.99 & 3.59 & 2.64 & 3.07 \\
        GPT-5.6 Sol & \textbf{4.08} & 3.80 & \textbf{4.96} & \textbf{4.28} & 3.93 & 3.56 & \textbf{4.71} & \textbf{4.06} & \textbf{3.60} & \textbf{3.96} & \textbf{4.51} & \textbf{4.02} \\
        \bottomrule
    \end{tabular}%
    }
\end{table*}

\begin{figure}[t]
    \centering
    \begin{minipage}[t]{0.49\linewidth}
        \vspace{-5pt}
        \centering
        \input{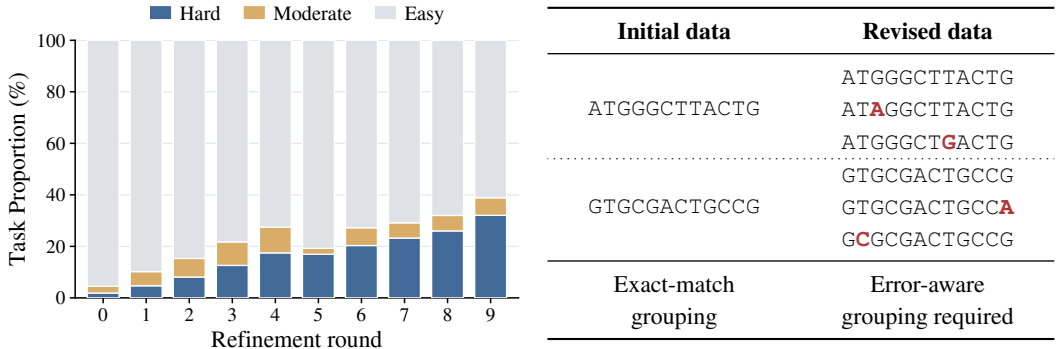}
    \end{minipage}
    \hfill
    \begin{minipage}[t]{0.48\linewidth}
        \vspace{0pt}
        \centering
        \begingroup
\small
\setlength{\tabcolsep}{0pt}
\renewcommand{\arraystretch}{1.21}

\setlength{\dashlinedash}{0.6pt}
\setlength{\dashlinegap}{2pt}

\definecolor{umierror}{RGB}{174,55,58}
\newcommand{\umiedit}[1]{\textcolor{umierror}{\textbf{#1}}}

\begin{tabular}{
    @{}
    >{\centering\arraybackslash}p{0.5\linewidth}
    >{\centering\arraybackslash}p{0.5\linewidth}
    @{}
}
    \toprule
    \textbf{Initial data} & \textbf{Revised data} \\
    \midrule

    \multirow{3}{*}{\texttt{ATGGGCTTACTG}}
    & \texttt{ATGGGCTTACTG} \\
    & \texttt{AT\umiedit{A}GGCTTACTG} \\
    & \texttt{ATGGGCT\umiedit{G}ACTG} \\

    \hdashline

    \multirow{3}{*}{\texttt{GTGCGACTGCCG}}
    & \texttt{GTGCGACTGCCG} \\
    & \texttt{GTGCGACTGCC\umiedit{A}} \\
    & \texttt{G\umiedit{C}GCGACTGCCG} \\

    \midrule

    Exact-match & Error-aware \\
    grouping & grouping required  \\

    \bottomrule
\end{tabular}

\endgroup
    \end{minipage}
    \vspace{-0.23in}
    \caption{(Left) Difficulty of generated tasks across refinement rounds. We report the proportion of tasks in each difficulty category at each refinement round.
    (Right) Actual data revision by iterative refinement.     Iterative refinement introduces sequencing errors that require error-aware grouping. The initial barcodes are error-free, whereas the revised barcodes contain altered bases, highlighted in red, that mimic errors encountered in real sequencing data.}
    \label{fig:iterative_refinement}
\end{figure}

\paragraph{Difficulty across Feedback-guided Refinement Rounds}
Fig.~\ref{fig:iterative_refinement} (Left) shows how the difficulty distribution changes across refinement rounds, pooling tasks generated by Claude-Opus-4.8 and GPT-5.6-Sol. The initial tasks are predominantly easy, indicating that the generating agent can solve most initial implementations. Over subsequent refinement rounds, the proportion of hard tasks increases, showing that solver feedback can be used to progressively construct tasks that expose failures not apparent in the initial generation. Difficulty over refinement rounds for each generator model is provided in Appendix~\ref{app:refinement_by_generator}.

\vspace{-0.01in}
\paragraph{Changes in Required Scientific Reasoning}
We further examine how iterative refinement changes the scientific reasoning required to solve a task.
As shown in Fig.~\ref{fig:iterative_refinement} (Right), the initial data provide evidence that can be interpreted directly from an intermediate result, whereas the revised task introduces a source of error in the data that requires the agent to revisit the raw data to interpret and validate the intermediate result.  Specifically, counting distinct unique molecular identifier (UMI) sequences recovers the number of original UMIs in the initial task, whereas the revision introduces UMI sequence errors, reflecting a source of inaccurate molecular counts in real sequencing workflows~\citep{smith2017umitools,sun2024correcting}. The revised task therefore requires agents to distinguish error-induced sequence variation from differences between original UMIs together with read support, as in UMI error-correction methods~\citep{yu2025afumic,piliper2026rumina}. Across all three solver trials on the revised task, the agents treated distinct UMI sequences as separate original UMIs, overlooking the error-containing sequencing, resulting in an incorrect answer. 
This example illustrates how iterative refinement can shift a task from direct use of an intermediate result toward interpretation of measurement artifacts and integration of multiple sources of evidence.

\vspace{-0.03in}
\subsection{Effect of Experience-guided Concept Generation}
\label{sec:experience_guided_generation_results}

\begin{wraptable}{r}{0.3\textwidth}
    \centering
    \vspace{-0.19in}
    \caption{Solver success rates and hard task proportions.}
    \label{tab:experience_difficulty}
    \vspace{-0.07in}
    \small
    \setlength{\tabcolsep}{5pt}
    \renewcommand{\arraystretch}{1.40}

    \begin{tabular}{@{}lcc@{}}
        \toprule
         & Success $\downarrow$ & Hard $\uparrow$ \\
        \midrule
        w/o Exp. & 66.7 & 15.0 \\
        w/ Exp.    & \textbf{35.0} & \textbf{50.0} \\
        \bottomrule
    \end{tabular}
    \vspace{-0.1in}
\end{wraptable}

\vspace{-0.03in}
\paragraph{Ablation Study on Experience-guided Concept Generation}
To see the effect of the experience guidance for concept generation, we compare tasks generated with and without experience-guided concept generation in materials science using Claude Opus 4.8. Both share 10 initial concepts and generate 50 tasks in total, retaining iterative task refinement. As shown in Table~\ref{tab:experience_difficulty}, experience-guided generation reduces solver success from 66.7\% to 35.0\% and increases the proportion of hard tasks from 15.0\% to 50.0\%. These results suggest that reusing prior refinement experience helps generate concepts that yield more challenging tasks than generating each concept independently.

\paragraph{Reuse of Required Solving Behaviors}
We further analyze how experience informs subsequent concept generation. As shown in Table~\ref{tab:experience_guided_concept_example} in Appendix~\ref{app:experience_concept_examples}, experience-guided concepts introduce a new domain, modality, and measurement, while drawing on prior experience to require similar solving behaviors. Specifically, experience from an earlier concept emphasizes identifying the appropriate analytical approach from observations, recovering the target quantity, and checking consistency with the data. A subsequently generated concept requires different scientific knowledge and analytical skills by introducing a different domain and modality while preserving these higher-level solving behaviors. The resulting task from the subsequent concept was rated as hard, whereas the task from the earlier concept was rated as moderate. This result suggests that experience can provide reusable guidance for designing new scientific tasks around higher-level reasoning behaviors that transfer across domains and modalities.

\subsection{Effect of Concept and Recipe Planning}
\label{sec:concept_recipe_ablation}

\begin{wrapfigure}{r}{0.37\textwidth}
    \vspace{-0.2in}
    \centering

    \includegraphics[width=\linewidth]{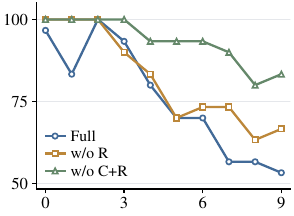}
    \vspace{-0.3in}
    \caption{Solver success (y-axis) across refinement rounds (x-axis).}
    \vspace{-0.1in}
    \label{fig:concept_recipe_ablation}
\end{wrapfigure}

To assess the contribution of concept and recipe planning to iterative refinement, we compare the full framework with two variants that remove recipe (w/o R) or both concept and recipe (w/o C+R) in materials science using Claude Opus 4.8. As shown in Fig.~\ref{fig:concept_recipe_ablation}, solver success decreases more substantially across refinement rounds in the full framework, while remaining higher when recipe planning or both stages are removed. Concepts and recipes specify the analytical requirements and the question and data design, respectively, providing explicit intermediate representations that can be revised based on solver feedback. Therefore, these results suggest that refining these structured plans is more effective at increasing task difficulty than directly revising the instantiated tasks.

\subsection{Qualitative Analyses on Generated tasks}

We further analyze representative generated tasks and their associated data files in Appendices~\ref{app:qa_example_msi} and \ref{app:generated_qa_examples}. The examples span diverse scientific topics, data modalities, and analytical requirements across all three domains. Rather than being answerable from question text or internal knowledge alone, these tasks require agents to interrogate scientific data, interpret intermediate observations, and integrate multiple sources of evidence to derive the final answer. Together, these examples show how AutoSciBench translates abstract evaluation requirements into concrete scientific tasks that exercise a broad range of agentic and scientific-reasoning capabilities, complementing the aggregate difficulty and quality analyses.
\section{Conclusion}
In this work, we investigated whether scientific-agent benchmarks can be automatically constructed and adapted as agent capabilities evolve. We developed AutoSciBench to study this question through iterative benchmark construction, solver-based evaluation, refinement, and experience-guided generation.
We designed concepts and recipes as high-level and low-level plans to define analytical requirements and guide the construction of questions and data. We further introduced feedback-guided refinement of these plans and experience-guided generation of subsequent concepts. Through experiments across three scientific domains, we demonstrate that AutoSciBench can generate benchmarks which expose solver failures and receive higher average quality scores than the original benchmarks. Further analyses suggest that iterative refinement increases task difficulty and can shift tasks toward data re-examination and interpretation, while prior experience guides the generation of challenging concepts. Together, results suggest that scientific-agent evaluation can itself become adaptive: solver behavior can guide the refinement of existing tasks, while experience accumulated across refinement trajectories can inform the generation of new evaluation targets as agent capabilities evolve.

\bibliography{iclr2027_conference}
\bibliographystyle{iclr2027_conference}

\newpage
\appendix
\section{Benchmark Generation Details}
\label{app:experimental_details}

\subsection{Seed Benchmarks}

\paragraph{CompBioBench}
CompBioBench~\citep{nair2026compbiobench} comprises 100 agentic computational-biology tasks spanning genomics, transcriptomics, epigenomics, single-cell analysis, human genetics, and machine-learning workflows. The tasks require multi-step reasoning, code execution, tool use, and interaction with external scientific resources.

\paragraph{MatTools}
MatTools~\citep{liu2025mattools} evaluates the ability to understand and use computational materials-science tools. It contains large-scale tasks derived from the \texttt{pymatgen} codebase and documentation, as well as realistic tool-use tasks requiring executable materials-property calculations.

\paragraph{MedCTA}
MedCTA~\citep{ashraf2026medcta} evaluates clinical tool agents on clinician-validated tasks grounded in multimodal inputs, including radiology images, pathology slides, and clinical reports. Its step-implicit questions require agents to determine how to acquire and integrate evidence without being given the intended tool sequence. This benchmark leverages the LLM evaluator to score the quality of answers, as the questions require open-ended answers.

\subsection{Concept Inference from Existing Benchmarks}

We use all 100 questions from CompBioBench and sample 100 questions from the code-generation tasks of MatTools and the training split of MedCTA. Each seed question is represented by its question text and paths to associated files. We process each seed question individually and instruct the agent to identify its scientific domain, data modality, and intended analytical requirements (i.e., operations). The agent returns a structured concept with a brief rationale, describing the evaluation target rather than a specific solution procedure. The prompt is provided in Fig.~\ref{fig:prompt_benchmark_concept_extraction}.

\subsection{Literature-Derived Workflows and Patterns}

We instruct the agent to search for scientific publications within the target field, prioritizing papers and preprints from the preceding five years. From each source, the agent extracts a workflow describing the input data, analytical steps, intermediate checks, and target outputs, together with the source reference. We then provide the extracted workflows to the agent to identify recurring analytical patterns, including shared analysis steps and decision points. These patterns abstract away study-specific details and tool choices, and their number is determined by the recurring structures supported by the collected workflows rather than a predefined count. The prompts for workflow discovery and pattern extraction are provided in Fig.~\ref{fig:prompt_workflow_discovery} and~\ref{fig:prompt_pattern_extraction}, respectively.

\subsection{Initial Concept Generation}

We provide the agent with existing benchmark-derived concepts, literature-derived workflows, and analytical patterns to generate initial concepts within the target field. The agent is instructed to generate a scientific domain, a data modality, and required operations along with a brief rationale, describing what a future task should test rather than prescribing a solution procedure. We generate five concepts per batch, providing previously proposed concepts to discourage duplicates and minor variations. The prompt is provided in Fig.~\ref{fig:prompt_initial_concept_generation}.

\subsection{Recipe Planning}

Given a concept, we instruct the agent to generate a recipe comprising seven fields: (i) data strategy which specifies whether to reuse existing benchmark data, retrieve data from the web, or simulate data, (ii) data plan which describes the required data and its answer-relevant information, (iii) construction mechanism which defines how the answer is established and concealed, (iv) ground-truth plan which specifies how to compute and verify the answer, (v) answer specification which defines the answer type and format, (vi) tools and resources which lists the required software and resources, and (vii) question-hiding strategy which specifies how to phrase the question without revealing solution steps. For clinical imaging, we encourage the agent to reuse images from existing benchmarks or retrieve real medical images from the web rather than generate simulated images. The prompt is provided in Fig.~\ref{fig:prompt_recipe_planning}.

\subsection{Task Implementation}

We provide the agent with a concept and its recipe and instruct it to construct a question, associated data files, and a reference answer. The agent acquires or generates the required data and executes code to compute and verify the answer. To ensure consistent scoring, we design each task to have a unique answer in a specified format, enabling exact-match evaluation. The question specifies the analysis target, input files, and required answer format without revealing the solution procedure. The prompt is provided in Fig.~\ref{fig:prompt_qa_implementation}.

\subsection{Solving and Judging}

We run three independent solver trials for each task, providing the question and associated data files without the reference answer. We evaluate solver answers using exact match against the reference answer. The judge reviews the question, data, reference answer, solver answers, and solving trajectories and returns three fields: (i) verifiability which indicates whether the question and data support a unique answer, distinguishing ambiguity from incorrect or incomplete analyses by the solver, (ii) difficulty which categorizes solver success across valid trials, and (iii) reasoning which explains the verifiability decision and the solve count underlying the difficulty rating. For verifiable tasks, difficulty is categorized as easy, moderate, or hard when all, some, or none of the valid trials succeed, respectively. The solving and judging prompts are provided in Fig.~\ref{fig:prompt_solving} and~\ref{fig:prompt_judging}, respectively.

\subsection{Feedback-guided Refinement}

If a task is not both verifiable and hard, we use judge feedback to revise its recipe or concept. Recipe revision preserves the concept while updating the data construction, answer computation and verification, or question design. After four consecutive recipe revisions without meeting both criteria, the agent revises the concept based on recent refinement history and generates a new recipe. The revised concept and recipe are used to reconstruct the task, followed by solving and judging. We repeat this process for up to 10 refinement rounds. The recipe and concept revision prompts are provided in Fig.~\ref{fig:prompt_recipe_refinement} and~\ref{fig:prompt_concept_refinement}, respectively.

\subsection{Experience-guided Concept Generation}

After refinement, we instruct the agent to summarize each completed history into five fields: (i) experience summary which summarizes the main design lessons, (ii) useful design patterns which identifies promising task construction strategies, (iii) anti-patterns which identifies designs to avoid, (iv) selection guidance which suggests evaluation targets to prioritize, and (v) generation guidance which suggests how to formulate future concepts. The experience summary abstracts away task-specific answers and data details. For subsequent concept generation, we provide the summary, design patterns, anti-patterns, and generation guidance together with the source priors and previously generated concepts. The prompts are provided in Fig.~\ref{fig:prompt_experience_construction} and~\ref{fig:prompt_experience_guidance}.

\subsection{Generation and Inference Configuration}
\label{app:gen_infer_configuration}

We implement the framework using Claude Code CLI with Claude Opus 4.8 or Codex CLI with GPT-5.6 Sol, using the same model across all stages within each generation condition. Each stage uses role-specific instructions and may involve multiple model turns and tool calls. For experience-guided concept generation, we provide up to 12 experience cards and 120 previously generated concepts, sampled uniformly at random under a fixed seed when the available records exceed these limits. We apply a timeout of 3,600 seconds to each pipeline stage. We set reasoning effort to high for all stages and leave temperature and token limits at their CLI defaults.

\section{Additional Experimental Results}
\label{app:additional_results}

\subsection{Additional Results on Quality Assessment}
\label{app:additional_quality}

\paragraph{Evaluator-wise Results}
We report quality scores separately for Claude Opus 4.8 and GPT-5.6 Sol in Tables~\ref{tab:quality_claude_judge} and~\ref{tab:quality_codex_judge}. The trends in scores assigned by both judges are consistent with the overall results in Table~\ref{tab:quality_overall}: generated benchmarks receive higher average scores than the human-curated benchmarks across all domains, with GPT-5.6-Sol-generated benchmarks achieving the highest average and fitness scores. Opus-4.8-generated benchmarks receive higher usefulness scores than GPT-5.6-Sol-generated benchmarks in computational biology and materials science under both judges. This consistency suggests that the observed quality differences are not driven solely by the preferences of a single judge model.

\begin{table*}[!h]
    \caption{Benchmark quality assessed by Claude Opus 4.8. Avg. denotes the mean of three criteria.}
    \centering
    \vspace{-0.1in}
    \label{tab:quality_claude_judge}
    \setlength{\tabcolsep}{4pt}
    \renewcommand{\arraystretch}{1.2}
    \resizebox{\textwidth}{!}{%
    \begin{tabular}{lcccacccaccca}
        \toprule
        & \multicolumn{4}{c}{Computational Biology} & \multicolumn{4}{c}{Materials Science} & \multicolumn{4}{c}{Clinical Imaging} \\
        \cmidrule(lr){2-5}\cmidrule(lr){6-9}\cmidrule(lr){10-13}
        & Verif. & Useful. & Fit. & Avg. & Verif. & Useful. & Fit. & Avg. & Verif. & Useful. & Fit. & Avg. \\
        \midrule
        Human & 2.72 & 3.56 & 3.45 & 3.24 & \textbf{4.58} & 1.13 & 1.46 & 2.39 & 2.01 & 2.21 & 1.45 & 1.89 \\
        \cdashline{1-13}[2pt/1pt]\rule{-2pt}{2.5ex}
        Opus 4.8 & 3.47 & \textbf{4.38} & 4.40 & 4.09 & 3.80 & \textbf{4.22} & 3.68 & 3.90 & 3.02 & 3.38 & 2.17 & 2.86 \\
        GPT-5.6 Sol & \textbf{4.02} & 3.68 & \textbf{4.92} & \textbf{4.21} & 3.90 & 3.40 & \textbf{4.62} & \textbf{3.97} & \textbf{3.39} & \textbf{3.78} & \textbf{4.26} & \textbf{3.81} \\
        \bottomrule
    \end{tabular}%
    }
\end{table*}

\begin{table*}[!h]
    \caption{Benchmark quality assessed by GPT-5.6 Sol. Avg. denotes the mean of the three criteria.}
    \centering
    \vspace{-0.1in}
    \label{tab:quality_codex_judge}
    \setlength{\tabcolsep}{4pt}
    \renewcommand{\arraystretch}{1.2}
    \resizebox{\textwidth}{!}{%
    \begin{tabular}{lcccacccaccca}
        \toprule
        & \multicolumn{4}{c}{Computational Biology} & \multicolumn{4}{c}{Materials Science} & \multicolumn{4}{c}{Clinical Imaging} \\
        \cmidrule(lr){2-5}\cmidrule(lr){6-9}\cmidrule(lr){10-13}
        & Verif. & Useful. & Fit. & Avg. & Verif. & Useful. & Fit. & Avg. & Verif. & Useful. & Fit. & Avg. \\
        \midrule
        Human & 2.18 & 3.68 & 3.54 & 3.13 & \textbf{4.94} & 1.84 & 1.32 & 2.70 & 2.74 & 2.96 & 2.00 & 2.57 \\
        \cdashline{1-13}[2pt/1pt]\rule{-2pt}{2.5ex}
        Opus 4.8 & 2.71 & \textbf{4.17} & 4.44 & 3.77 & 3.18 & \textbf{4.12} & 4.24 & 3.85 & 2.95 & 3.80 & 3.09 & 3.28 \\
        GPT-5.6 Sol & \textbf{4.14} & 3.92 & \textbf{5.00} & \textbf{4.35} & 3.96 & 3.71 & \textbf{4.80} & \textbf{4.16} & \textbf{3.81} & \textbf{4.14} & \textbf{4.77} & \textbf{4.24} \\
        \bottomrule
    \end{tabular}%
    }
\end{table*}

\paragraph{Inter-Judge Agreement}
We further evaluate the agreement between the two judges. For each criterion, we compute Spearman's $\rho$ and Kendall's $\tau_b$ between the two judges' scores to measure whether they order tasks similarly. As shown in Table~\ref{tab:quality_judge_agreement}, both metrics show positive correlations across all criteria, indicating that the two judges generally agree on the relative quality of tasks. These results support the consistency of our relative task-quality assessments across the two judge models, providing robustness of the quality evaluation reported in Table~\ref{tab:quality_overall} in the main paper.

\begin{table}[!h]
    \centering
    \caption{Inter-judge agreement measured by Spearman's $\rho$ and Kendall's $\tau_b$ between the two judges.}
    \vspace{-0.05in}
    \label{tab:quality_judge_agreement}
    \small
    \setlength{\tabcolsep}{5pt}
    \renewcommand{\arraystretch}{1.2}
    \begin{tabular}{lcccccc}
        \toprule
        & \multicolumn{2}{c}{Verifiability}
        & \multicolumn{2}{c}{Usefulness}
        & \multicolumn{2}{c}{Fitness} \\
        \cmidrule(lr){2-3}\cmidrule(lr){4-5}\cmidrule(lr){6-7}
        & $\rho$ & $\tau_b$ & $\rho$ & $\tau_b$ & $\rho$ & $\tau_b$ \\
        \midrule
        CompBioBench & 0.67 & 0.59 & 0.59 & 0.54 & 0.73 & 0.68 \\
        MatTools & 0.57 & 0.53 & 0.90 & 0.81 & 0.90 & 0.83 \\
        MedCTA & 0.75 & 0.67 & 0.82 & 0.75 & 0.81 & 0.75 \\
        \midrule
        All benchmarks & 0.74 & 0.64 & 0.84 & 0.77 & 0.87 & 0.79 \\
        \bottomrule
    \end{tabular}
\end{table}

\subsection{Effect of Feedback-guided Iterative Refinement by Generation Model}
\label{app:refinement_by_generator}

\begin{figure}[t]
    \centering
    \begin{minipage}[t]{0.485\linewidth}
        \centering
        {\phantom{text}\textbf{Claude-Opus-4.8}\par}
        \smallskip
        \includegraphics[width=\linewidth]{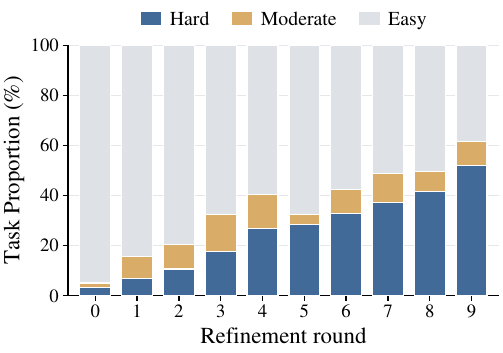}
    \end{minipage}
    \hfill
    \begin{minipage}[t]{0.485\linewidth}
        \centering
        {\phantom{text}\textbf{GPT-5.6-Sol}\par}
        \smallskip
        \includegraphics[width=\linewidth]{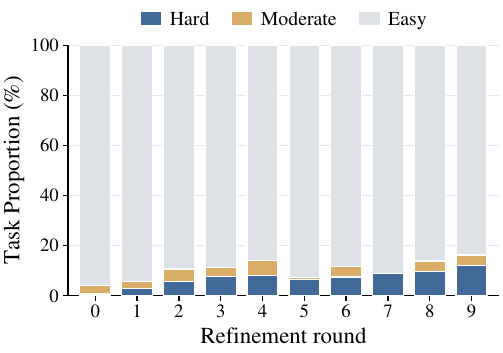}
    \end{minipage}
    \vspace{-0.1in}
    \caption{Difficulty across refinement rounds for tasks generated by Claude-Opus-4.8 (Left) and GPT-5.6-Sol (Right). We report proportions by gathering tasks across all three benchmarks.}
    \label{fig:refinement_by_generator}
\end{figure}

We separately analyze the difficulty of tasks generated by Claude-Opus-4.8 and GPT-5.6-Sol across refinement rounds. As shown in Fig.~\ref{fig:refinement_by_generator}, the proportion of hard tasks increases for both models, with a larger increase for Claude-Opus-4.8. This result suggests that Claude-Opus-4.8 is particularly useful for constructing tasks that challenge its own solver through iterative refinement.

\subsection{Concept Examples for Experience-guided Generation}
\label{app:experience_concept_examples}

Table~\ref{tab:experience_guided_concept_example} compares an earlier concept after refinement with a subsequently generated concept. The required measurements and physical models are different, whereas the overall analytical structure is similar, reflecting the reasoning and tool-use patterns of the earlier concept into the new concept. Specifically, the required operations in both concepts involve (i) identification, (ii) recovery, and  (iii) consistency checking, while the target quantities and data modalities differ. Therefore, the new concept adopts the analytical structure of the earlier concept, while changing the specific measurements and models to create a new evaluation target.

\begin{table*}[t!]
    \centering
    \definecolor{experienceModel}{HTML}{27649D}
    \definecolor{experienceRecovery}{HTML}{A65318}
    \definecolor{experienceConsistency}{HTML}{18736A}
    \caption{Full concept descriptions for the experience-guided generation example. The left column shows the previous concept, and the right column shows a concept generated with the experience. Matching colors indicate corresponding solving behaviors: model identification (blue), quantity recovery (orange), and consistency checking (teal).}
    \label{tab:experience_guided_concept_example}

    \small
    \setlength{\tabcolsep}{0pt}
    \renewcommand{\arraystretch}{1.15}

    \begin{tabular}{
        @{}
        >{\raggedright\arraybackslash}p{0.12\textwidth}
        @{\hspace{0.025\textwidth}}
        >{\raggedright\arraybackslash}p{0.415\textwidth}
        @{\hspace{0.025\textwidth}}
        >{\raggedright\arraybackslash}p{0.415\textwidth}
        @{}
    }
        \toprule
        & \textbf{Earlier concept}
        & \textbf{Subsequent concept} \\
        \midrule

        Domain
        & semiconductor junction electrostatics and doping-profile characterization
        & semiconductor optical spectroscopy / band-gap determination \\

        \midrule
        Modality
        & a measured junction electrical characteristic (capacitance-versus-reverse-bias profile, optionally with temperature or a second characteristic) for a junction whose doping profile and built-in potential are undisclosed
        & a tabulated optical absorption coefficient versus photon energy spanning the absorption edge of a semiconductor thin film, with no transition-type label supplied \\

        \midrule

        Operations
        & \vspace{-0.6\baselineskip}
        \begin{itemize}[leftmargin=*,topsep=0pt,itemsep=2pt,partopsep=0pt,parsep=0pt]
              \item \textcolor{experienceModel}{identification of the junction doping-profile regime (abrupt versus graded)} from the functional dependence of junction capacitance on applied reverse bias
              \item \textcolor{experienceRecovery}{recovery of the latent built-in potential and effective doping} consistent with that regime from the measured characteristic
              \item \textcolor{experienceConsistency}{reconciliation of the depletion-derived estimate with the profile regime it implies} into a single self-consistent value
          \end{itemize}
        & \vspace{-0.6\baselineskip}
        \begin{itemize}[leftmargin=*,topsep=0pt,itemsep=2pt,partopsep=0pt,parsep=0pt]
              \item \textcolor{experienceModel}{discrimination between a direct-allowed and an indirect-allowed interband transition} by which tauc exponent linearizes the $(\mathrm{alpha}\cdot h\cdot\mathrm{nu})^{(1/n)}$ versus photon-energy edge
              \item \textcolor{experienceRecovery}{recovery of the optical band-gap energy} as the linear-extrapolation intercept of the correctly linearized tauc edge
              \item \textcolor{experienceConsistency}{rejection of the incorrect exponent that yields a curved (non-linear) edge}
          \end{itemize} \\

        \bottomrule
    \end{tabular}
\end{table*}

\clearpage
\subsection{Qualitative Analysis of the Microsatellite Instability Task}
\label{app:qa_example_msi}

\begingroup
\tcbset{msiexample/.style={
    enhanced,breakable,colback=white,colframe=black!35,
    colbacktitle=black!5,coltitle=black,fonttitle=\small\bfseries,
    fontupper=\small,boxrule=0.5pt,arc=1mm,
    left=8pt,right=8pt,top=6pt,bottom=6pt,before skip=7pt,after skip=7pt
}}
\lstdefinestyle{msidata}{language={},basicstyle=\ttfamily\fontsize{8}{9.5}\selectfont,
    columns=fullflexible,keepspaces=true,showstringspaces=false,
    breaklines=true,aboveskip=0pt,belowskip=0pt}

\begin{tcolorbox}[msiexample,title={Question},before upper={\raggedright}]
You are given whole-panel sequencing data for one hypothetical cancer case: a matched tumor/normal pair aligned to a synthetic reference at a panel of 45 mononucleotide microsatellite marker loci. The tumor is an admixture of tumor and non-tumor cells, and its tumor purity is not provided anywhere in the data.
\par\smallskip
Input files (all in the current directory):
\begin{itemize}[leftmargin=12pt,nosep,topsep=3pt]
    \item \texttt{reference.fa} (and its index \texttt{reference.fa.fai}): the reference; each marker is its own contig containing a homopolymer tract embedded in unique flanking sequence.
    \item \texttt{panel.tsv}: one row per marker with columns \texttt{marker\_id}, \texttt{contig}, \texttt{tract\_start} (0-based, inclusive), \texttt{tract\_end} (0-based, exclusive), \texttt{reference\_repeat\_length}, \texttt{repeat\_unit}.
    \item \texttt{normal.bam} (and index \texttt{normal.bam.bai}): matched-normal alignments.
    \item \texttt{tumor.bam} (and index \texttt{tumor.bam.bai}): tumor alignments.
\end{itemize}
\smallskip
Using only these files, determine how many of the 45 panel markers exhibit genuine somatic microsatellite instability in the tumor (i.e., tumor-acquired repeat-length instability that is not explained by the individual's germline or by the matched normal). Report your answer in the exact format specified below.
\par\smallskip
Answer format: a single integer rendered as a plain base-10 string with no leading zeros, no sign, no spaces, and no thousands separators (valid range 0--45), e.g. ``7''.
\end{tcolorbox}

\begin{tcolorbox}[msiexample,title={Provided Data (Excerpts)},before upper={\raggedright}]
\textbf{\texttt{reference.fa}}\par\nobreak\smallskip\nobreak
\begin{lstlisting}[style=msidata]
>MSM27
TGGCTTTCCTGGTTCTCGGGCTCGTTCTTTTCGGTCGTTCTTTCGGGTTTAAAAAAAAAAAAAAAAAAAA
AAAAAAAAAAAAAAAAAGTCCGGGGCCGGTTGGCGGGCTCGTCGGGTTGGCCTTTTTTGCCTTGGGC
\end{lstlisting}
\medskip
\textbf{\texttt{reference.fa.fai}}\par\nobreak\smallskip\nobreak
{\ttfamily\footnotesize
\setlength{\tabcolsep}{5pt}
\renewcommand{\arraystretch}{1.12}
\begin{tabular}{@{}lllll@{}}
\toprule
contig & length & offset & bases/line & bytes/line \\
\midrule
MSM27 & 137 & 3711 & 70 & 71 \\
\bottomrule
\end{tabular}\par}
\medskip
\textbf{\texttt{panel.tsv}}\par\nobreak\smallskip\nobreak
{\ttfamily\fontsize{8}{9.5}\selectfont
\setlength{\tabcolsep}{3pt}
\renewcommand{\arraystretch}{1.12}
\begin{tabular}{@{}llllll@{}}
\toprule
marker\_id & contig & tract\_start & tract\_end & reference\_repeat\_length & repeat\_unit \\
\midrule
MSM04 & MSM04 & 50 & 86 & 36 & A \\
MSM15 & MSM15 & 50 & 86 & 36 & A \\
MSM16 & MSM16 & 50 & 87 & 37 & A \\
MSM18 & MSM18 & 50 & 86 & 36 & A \\
MSM27 & MSM27 & 50 & 87 & 37 & A \\
MSM28 & MSM28 & 50 & 80 & 30 & A \\
\bottomrule
\end{tabular}\par}
\medskip
\textbf{\texttt{normal.bam}}\par\nobreak\smallskip\nobreak
{\ttfamily\footnotesize
\setlength{\tabcolsep}{5pt}
\renewcommand{\arraystretch}{1.12}
\begin{tabular}{@{}llllll@{}}
\toprule
QNAME & FLAG & RNAME & POS & MAPQ & CIGAR \\
\midrule
MSM27\_normal\_0 & 0 & MSM27 & 1 & 60 & 137M \\
MSM27\_normal\_1400 & 0 & MSM27 & 1 & 60 & 85M2D50M \\
\bottomrule
\end{tabular}\par}
\medskip
\textbf{\texttt{tumor.bam}}\par\nobreak\smallskip\nobreak
{\ttfamily\footnotesize
\setlength{\tabcolsep}{5pt}
\renewcommand{\arraystretch}{1.12}
\begin{tabular}{@{}llllll@{}}
\toprule
QNAME & FLAG & RNAME & POS & MAPQ & CIGAR \\
\midrule
MSM27\_tumor\_0 & 0 & MSM27 & 1 & 60 & 137M \\
MSM27\_tumor\_1280 & 0 & MSM27 & 1 & 60 & 85M2D50M \\
\bottomrule
\end{tabular}\par}
\medskip
\textbf{\texttt{normal.bam.bai} and \texttt{tumor.bam.bai}}\par\nobreak\smallskip\nobreak
{\ttfamily\footnotesize
\setlength{\tabcolsep}{5pt}
\renewcommand{\arraystretch}{1.12}
\begin{tabular}{@{}ll@{}}
\toprule
file & first 8 bytes (hex) \\
\midrule
normal.bam.bai & 42 41 49 01 2d 00 00 00 \\
tumor.bam.bai & 42 41 49 01 2d 00 00 00 \\
\bottomrule
\end{tabular}\par}
\end{tcolorbox}

\begin{tcolorbox}[msiexample,title={Ground-Truth Answer}]
\texttt{13}
\end{tcolorbox}
\endgroup

\begin{figure}[htbp]
    \centering
    \includegraphics[width=\linewidth]{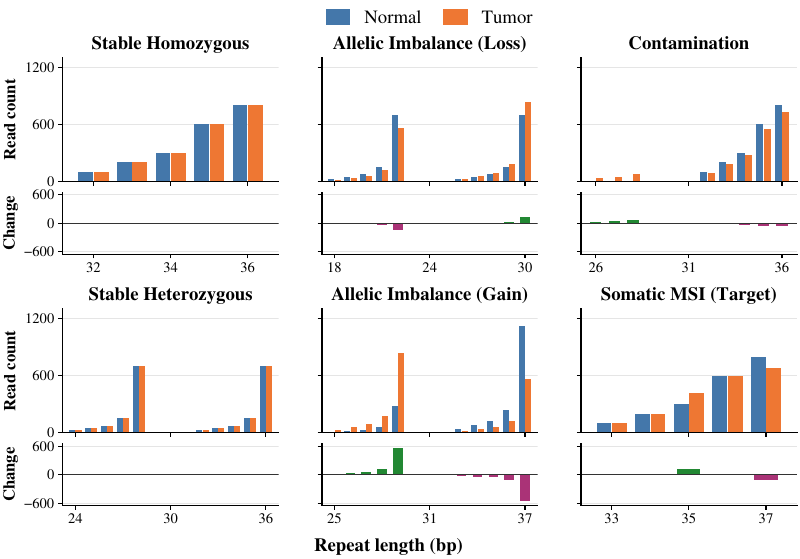}
    \caption{Representative marker distributions for the MSI tasks. Each panel shows normal and tumor read counts (upper) and their difference, tumor minus normal (lower), extracted from the BAM files. Each sample contains 2,000 reads per marker. Contamination introduces new repeat lengths, whereas the target example redistributes reads among existing lengths. Class labels describe the synthetic construction and are not supplied to the solver.}
    \label{fig:msi_example}
\end{figure}

In this section, we provide an in-depth analysis of a generated task. As shown in Question and Data files above, it asks agents to determine the number of markers with genuine somatic microsatellite instability (MSI) from paired tumor and normal sequencing data. The input consists of a reference sequence, marker coordinates, and two BAM files containing aligned sequencing reads. Each sequencing read shows different characteristics. Specifically, as shown in Fig.~\ref{fig:msi_example}, there are six types of constructed sequencing data: stable homozygous and heterozygous markers, allelic imbalance (Loss), allelic imbalance (Gain), contamination, and somatic MSI. The solver should extract repeat-length distributions from these files and distinguish tumor-acquired instability based on these distributional characteristics. 

The analysis requires using Python with \texttt{pysam} to read the reference sequence and BAM alignments. First, the solver should use \texttt{panel.tsv} and \texttt{reference.fa} to locate each repeat tract and identify its flanking sequences. It should then extract reads spanning the tract from \texttt{normal.bam} and \texttt{tumor.bam}, considering insertions and deletions when recovering repeat lengths. Then, the solver should aggregate these measurements into read counts and relative frequencies for each repeat length, producing paired distributions for all 45 markers. Therefore, these steps require tool implementation and execution, along with checks that the extracted measurements correspond to the intended repeat tract rather than the full read or reference length.

The key to this task is interpreting the pattern of distributional changes rather than simply detecting differences between tumor and normal samples. In the illustrated somatic MSI example, short deletions within the repeat tract shift read counts toward shorter repeat lengths. Distinguishing this pattern from allelic imbalance and contamination requires domain-specific knowledge and scientific interpretation beyond extracting the distributions.

\subsection{Analysis on Generation Cost}
\label{app:generation_cost}

We further analyze the generation cost per question and most consuming generation stage. As shown in Table~\ref{tab:cost_per_question}, the cost of Opus 4.8 is smaller than that of GPT-5.6 Sol across all three domains, while its generated benchmarks yield lower average solver accuracy. This result suggests that Opus 4.8 is more cost-effective for generating challenging tasks. Table~\ref{tab:cost_proportion} shows the most consuming stages in AutoSciBench. The task implementation and solving stages are the most consuming stages in both models, which is expected as these stages require the model to implement and execute the analysis.

\begin{table}[!htbp]
    \centering
    \caption{Cost per question (USD) on each domain and average benchmark difficulty in Table~\ref{tab:main_results}.}
    \label{tab:cost_per_question}
    \vspace{-0.1in}
    \small
    \setlength{\tabcolsep}{6pt}
    \renewcommand{\arraystretch}{1.3}
    \begin{tabular}{lcccc}
        \toprule
        & Computational Biology & Materials Science & Clinical Imaging & Avg. Solver Accuracy \\
        \midrule
        Opus 4.8 & \$58.47 & \$22.36 & \$24.88 & 66.2 \\
        GPT-5.6 Sol & \$111.26 & \$114.11 & \$110.45 & 75.4 \\
        \bottomrule
    \end{tabular}%
\end{table}

\begin{table}[!htbp]
    \centering
    \caption{Cost proportions (\%) for implementation (Impl.) and solving (Solve). Total denotes their combined proportion.}
    \label{tab:cost_proportion}
    \vspace{-0.05in}
    \small
    \setlength{\tabcolsep}{6pt}
    \renewcommand{\arraystretch}{1.3}
    \begin{tabular}{lcc>{\columncolor{black!8}}ccc>{\columncolor{black!8}}ccc>{\columncolor{black!8}}c}
        \toprule
        & \multicolumn{3}{c}{Computational Biology}
        & \multicolumn{3}{c}{Material Science}
        & \multicolumn{3}{c}{Clinical Imaging} \\
        \cmidrule(lr){2-4}\cmidrule(lr){5-7}\cmidrule(lr){8-10}
        & Impl. & Solve & Total
        & Impl. & Solve & Total
        & Impl. & Solve & Total \\
        \midrule
        Opus 4.8 & 47.5 & 44.7 & 92.3 & 46.2 & 41.2 & 87.4 & 51.3 & 28.0 & 79.3 \\
        GPT-5.6 Sol & 37.5 & 51.9 & 89.4 & 42.7 & 47.6 & 90.3 & 66.3 & 26.0 & 92.3 \\
        \bottomrule
    \end{tabular}%
\end{table}

\clearpage
\section{Examples of Generated Benchmarks}
\label{app:generated_qa_examples}

We present two examples for each domain and generation model, covering different scientific subdomains, data formats, and analytical requirements.
Questions and ground-truth answers are reproduced in full, while tabular data are shown as excerpts with selected rows and, where indicated, selected columns.

\subsection{Computational Biology: Claude Opus 4.8}
\label{app:qa_example_compbio_opus48}

\paragraph{Example 1: Antigen Presentation}
The question asks the solver to identify the candidate peptide with the highest predicted presentation on the MHC class I allele HLA-A*02:01.
The generator fetches protein sequences from UniProt~\citep{uniprot2025}\footnote{Data from the UniProt Consortium
(\url{https://www.uniprot.org/}), licensed under
\href{https://creativecommons.org/licenses/by/4.0/}{CC BY 4.0}.} and stores them in \texttt{source\_proteins.fasta}.
It extracts and selects 30 candidate peptides using MHCflurry predictions and constructs \texttt{candidate\_peptides.tsv}, recording each peptide's source protein accession and position while leaving its surrounding amino-acid context to be recovered from the FASTA file.

\begingroup
\tcbset{qaexample/.style={
    enhanced,
    breakable,
    colback=white,
    colframe=black!35,
    colbacktitle=black!5,
    coltitle=black,
    fonttitle=\small\bfseries,
    fontupper=\small,
    boxrule=0.5pt,
    arc=1mm,
    left=8pt,
    right=8pt,
    top=6pt,
    bottom=6pt,
    before skip=7pt,
    after skip=7pt
}}

\begin{tcolorbox}[qaexample,title={Question},before upper={\raggedright}]
You are given two files. \texttt{candidate\_peptides.tsv} is a tab-delimited table with a header row and exactly these columns, in order: \texttt{peptide\_id}, \texttt{uniprot\_id}, \texttt{start}, \texttt{peptide}. Each row is one candidate: \texttt{peptide\_id} is an identifier; \texttt{uniprot\_id} is a UniProt accession that is present in \texttt{source\_proteins.fasta}; \texttt{start} is the 0-based index at which \texttt{peptide} occurs within that protein's sequence; and \texttt{peptide} is the candidate amino-acid sequence in uppercase one-letter codes. \texttt{source\_proteins.fasta} is a standard FASTA file (one record per referenced accession, each header the bare accession) containing the proteins referenced in the table. Using MHCflurry 2.2.1 with its default built-in class I presentation predictor, identify the single candidate that is most likely to be presented on the MHC class I allele HLA-A*02:01. Return exactly one peptide sequence: the value from the \texttt{peptide} column for that candidate.
\end{tcolorbox}

\begin{tcolorbox}[qaexample,title={Provided Data (Excerpts)}]
\textbf{\texttt{candidate\_peptides.tsv}}\par\smallskip
{\raggedright
\setlength{\tabcolsep}{10pt}
\renewcommand{\arraystretch}{1.12}
\begin{tabular}{@{}llll@{}}
\toprule
\texttt{peptide\_id} & \texttt{uniprot\_id} & \texttt{start} & \texttt{peptide} \\
\midrule
\texttt{P12} & \texttt{P01375} & \texttt{194} & \texttt{YLGGVFQL} \\
\texttt{P13} & \texttt{P68871} & \texttt{13}  & \texttt{ALWGKVNV} \\
\texttt{P14} & \texttt{P01375} & \texttt{67}  & \texttt{SLISPLAQAV} \\
\texttt{P15} & \texttt{P0DTC2} & \texttt{252} & \texttt{DSSSGWTA} \\
\bottomrule
\end{tabular}\par}

\medskip
\textbf{\texttt{source\_proteins.fasta}}\par\smallskip
\begin{lstlisting}[language={},basicstyle=\ttfamily\footnotesize,columns=fullflexible,keepspaces=true,showstringspaces=false,breaklines=true,aboveskip=0pt,belowskip=0pt]
>P01375
MSTESMIRDVELAEEALPKKTGGPQGSRRCLFLSLFSFLIVAGATTLFCLLHFGVIGPQR
EEFPRDLSLISPLAQAVRSSSRTPSDKPVAHVVANPQAEGQLQWLNRRANALLANGVELR
DNQLVVPSEGLYLIYSQVLFKGQGCPSTHVLLTHTISRIAVSYQTKVNLLSAIKSPCQRE
TPEGAEAKPWYEPIYLGGVFQLEKGDRLSAEINRPDYLDFAESGQVYFGIIAL
\end{lstlisting}
\end{tcolorbox}

\begin{tcolorbox}[qaexample,title={Ground-Truth Answer}]
\texttt{SLISPLAQAV}
\end{tcolorbox}

\endgroup

\clearpage
\paragraph{Example 2: Gene Essentiality}
The question asks the solver to identify the strongest genuinely essential gene from a pooled CRISPR knockout screen.
The generator simulates guide-level read counts before and after selection in \texttt{readcounts.tsv} and constructs gene-level copy-number annotations in \texttt{copynumber.tsv}.
It incorporates copy-number-dependent depletion and non-functional guides so that the strongest observed depletion does not directly identify the most essential gene.

    \begingroup
\tcbset{qaexample/.style={
    enhanced,
    breakable,
    colback=white,
    colframe=black!35,
    colbacktitle=black!5,
    coltitle=black,
    fonttitle=\small\bfseries,
    fontupper=\small,
    boxrule=0.5pt,
    arc=1mm,
    left=8pt,
    right=8pt,
    top=6pt,
    bottom=6pt,
    before skip=7pt,
    after skip=7pt
}}

\begin{tcolorbox}[qaexample,title={Question},before upper={\raggedright}]
You are given the guide-level results of a pooled CRISPR knockout screen in two tab-separated files in the current working directory:

\begin{itemize}[leftmargin=*,nosep]
\item \texttt{readcounts.tsv} \textemdash{} one row per sgRNA guide, with columns: \texttt{guide\_id}, \texttt{gene\_id}, \texttt{plasmid\_T0} (plasmid/t0 reference read count), \texttt{endpoint\_rep1}, \texttt{endpoint\_rep2}, \texttt{endpoint\_rep3} (post-selection endpoint replicate read counts). All counts are integers. A block of non-targeting control guides is included (\texttt{gene\_id = CTRL}).
\item \texttt{copynumber.tsv} \textemdash{} one row per targeted gene, with columns: \texttt{gene\_id}, \texttt{locus} (genomic segment), \texttt{copy\_number} (integer segment copy number at that locus).
\end{itemize}

Using BOTH files, report the single gene that is the strongest genuinely essential gene in this screen \textemdash{} the gene whose knockout produces the strongest genuine dropout at the endpoint relative to the plasmid/t0 reference. Report exactly one \texttt{gene\_id}, exactly as it appears in the \texttt{gene\_id} column of the files.
\end{tcolorbox}

\begin{tcolorbox}[qaexample,title={Provided Data (Excerpts)},before upper={\raggedright}]
\textbf{\texttt{readcounts.tsv}}\par\smallskip
{\ttfamily\footnotesize
\setlength{\tabcolsep}{4pt}
\renewcommand{\arraystretch}{1.12}
\begin{tabular}{@{}llllll@{}}
\toprule
guide\_id & gene\_id & plasmid\_T0 & endpoint\_rep1 & endpoint\_rep2 & endpoint\_rep3 \\
\midrule
sg0001 & G015 & 1084 & 1049 & 1251 & 1661 \\
sg0002 & G015 & 888 & 843 & 843 & 840 \\
sg0003 & G015 & 841 & 25 & 28 & 23 \\
sg0004 & G015 & 903 & 37 & 44 & 28 \\
sg0005 & G015 & 1159 & 49 & 27 & 51 \\
sg0006 & G015 & 845 & 39 & 34 & 25 \\
\bottomrule
\end{tabular}\par}

\medskip
\textbf{\texttt{copynumber.tsv}}\par\smallskip
{\ttfamily\footnotesize
\setlength{\tabcolsep}{5pt}
\renewcommand{\arraystretch}{1.12}
\begin{tabular}{@{}lll@{}}
\toprule
gene\_id & locus & copy\_number \\
\midrule
G015 & chr17:165009552-165290996 & 10 \\
G016 & chr18:116393201-116622843 & 2 \\
G084 & chr3:44897813-45197745 & 20 \\
\bottomrule
\end{tabular}\par}
\end{tcolorbox}

\begin{tcolorbox}[qaexample,title={Ground-Truth Answer}]
\texttt{G015}
\end{tcolorbox}
\endgroup

\clearpage
\subsection{Computational Biology: GPT-5.6-Sol}
\label{app:qa_example_compbio_gpt56sol}

\paragraph{Example 1: Kinase-Substrate Identification}
The question asks the solver to identify the single directly phosphorylated kinase-substrate site in a synthetic phosphoproteomics case.
The generator simulates mass spectra, phosphopeptide time courses, and protein-abundance controls across treatments and replicates.
It constructs the accompanying peptide metadata, inhibitor-selectivity values, motif annotations, and calibration information so that the site can be identified by combining the simulated observations.

\begingroup
\tcbset{qaexample/.style={
    enhanced,
    breakable,
    colback=white,
    colframe=black!35,
    colbacktitle=black!5,
    coltitle=black,
    fonttitle=\small\bfseries,
    fontupper=\small,
    boxrule=0.5pt,
    arc=1mm,
    left=8pt,
    right=8pt,
    top=6pt,
    bottom=6pt,
    before skip=7pt,
    after skip=7pt
}}

\begin{tcolorbox}[qaexample,title={Question},before upper={\raggedright}]
Given \texttt{phospho\_spectra.tsv}, \texttt{peptide\_spectrum\_metadata.tsv}, \texttt{phosphopeptide\_timecourse.tsv}, \texttt{protein\_abundance\_controls.tsv}, \texttt{inhibitor\_selectivity.tsv}, \texttt{kinase\_motif\_annotations.tsv}, \texttt{calibration\_controls.json}, and \texttt{case\_schema.json}, identify the single directly phosphorylated kinase--substrate site represented by this synthetic case. Use the one-based canonical residue numbering, synthetic UniProt-style substrate accession namespace, kinase symbols, treatment labels, and time labels defined in \texttt{case\_schema.json}. Return only one ASCII string in the format \texttt{KINASE\_SYMBOL::\allowbreak SUBSTRATE\_ACCESSION::\allowbreak RESIDUE\_POSITION::\allowbreak DIRECT}, with no spaces or commentary.
\end{tcolorbox}

\begin{tcolorbox}[qaexample,title={Provided Data (Excerpts)}]
\textbf{\texttt{phospho\_spectra.tsv}}\par\smallskip
{\raggedright
\ttfamily\footnotesize
\setlength{\tabcolsep}{10pt}
\renewcommand{\arraystretch}{1.12}
\begin{tabular}{@{}lll@{}}
\toprule
spectrum\_id & mz & intensity \\
\midrule
SP001 & 114.090848 & 9071.62 \\
SP001 & 147.112193 & 9772.88 \\
SP001 & 151.165436 & 183.20 \\
\bottomrule
\end{tabular}\par}

\medskip
\textbf{\texttt{phosphopeptide\_timecourse.tsv}}\par\smallskip
{\raggedright
\ttfamily\footnotesize
\setlength{\tabcolsep}{5pt}
\renewcommand{\arraystretch}{1.12}
\begin{tabular}{@{}lllll@{}}
\toprule
feature\_id & treatment & time\_min & replicate & intensity \\
\midrule
F001 & UNSTIM & 2 & 1 & 10002.716 \\
F001 & STIM & 2 & 1 & 39594.723 \\
F001 & STIM\_MAPK1\_INH & 2 & 1 & 8080.134 \\
\bottomrule
\end{tabular}\par}

\medskip
\textbf{\texttt{protein\_abundance\_controls.tsv}}\par\smallskip
{\raggedright
\ttfamily\footnotesize
\setlength{\tabcolsep}{5pt}
\renewcommand{\arraystretch}{1.12}
\begin{tabular}{@{}lllll@{}}
\toprule
substrate\_accession & treatment & time\_min & replicate & intensity \\
\midrule
P0SYN1 & UNSTIM & 2 & 1 & 2000.586 \\
P0SYN1 & STIM & 2 & 1 & 2024.132 \\
P0SYN1 & STIM\_MAPK1\_INH & 2 & 1 & 2039.675 \\
\bottomrule
\end{tabular}\par}
\end{tcolorbox}

\begin{tcolorbox}[qaexample,title={Ground-Truth Answer}]
\texttt{MAPK1::P0SYN1::S383::DIRECT}
\end{tcolorbox}
\endgroup

\clearpage
\paragraph{Example 2: Metagenomic Contamination}
The question asks the solver to identify the contig that originates from a different genome than the majority of members in a metagenome-assembled genome.
The generator simulates contig sequences, coverage across samples, and paired-read links, and constructs a genome-membership table containing one contig from a different source genome.
It gives this contig a sequence composition resembling the majority genome while retaining coverage and linkage evidence that distinguishes its origin.

\begingroup
\tcbset{qaexample/.style={
    enhanced,
    breakable,
    colback=white,
    colframe=black!35,
    colbacktitle=black!5,
    coltitle=black,
    fonttitle=\small\bfseries,
    fontupper=\small,
    boxrule=0.5pt,
    arc=1mm,
    left=8pt,
    right=8pt,
    top=6pt,
    bottom=6pt,
    before skip=7pt,
    after skip=7pt
}}

\begin{tcolorbox}[qaexample,title={Question},before upper={\raggedright}]
Using \texttt{README.md}, \texttt{assembly.fasta}, \texttt{mag\_membership.tsv}, \texttt{coverage.tsv}, and \texttt{paired\_links.tsv}, determine the unique \texttt{mag\_01} contig that originates from a different genome than the strict majority of \texttt{mag\_01} members. Return exactly one case-sensitive identifier matching \texttt{ctg\_NNN}, with no whitespace or surrounding text.
\end{tcolorbox}

\begin{tcolorbox}[qaexample,title={Provided Data (Excerpts)},before upper={\raggedright}]
\textbf{\texttt{mag\_membership.tsv}}\par\smallskip
{\ttfamily\footnotesize
\setlength{\tabcolsep}{5pt}
\renewcommand{\arraystretch}{1.12}
\begin{tabular}{@{}ll@{}}
\toprule
mag\_id & contig\_id \\
\midrule
mag\_01 & ctg\_023 \\
mag\_01 & ctg\_025 \\
mag\_01 & ctg\_048 \\
mag\_01 & ctg\_049 \\
\bottomrule
\end{tabular}\par}

\medskip
\textbf{\texttt{coverage.tsv}} (selected columns)\par\smallskip
{\ttfamily\footnotesize
\setlength{\tabcolsep}{5pt}
\renewcommand{\arraystretch}{1.12}
\begin{tabular}{@{}lll@{}}
\toprule
contig\_id & sample\_id & coverage\_ppm \\
\midrule
ctg\_001 & q01\_x0y0 & 1734 \\
ctg\_001 & q01\_x0y1 & 1451 \\
ctg\_001 & q01\_x1y0 & 2018 \\
ctg\_001 & q01\_x1y1 & 1746 \\
\bottomrule
\end{tabular}\par}

\medskip
\textbf{\texttt{paired\_links.tsv}} (selected columns)\par\smallskip
{\ttfamily\footnotesize
\setlength{\tabcolsep}{5pt}
\renewcommand{\arraystretch}{1.12}
\begin{tabular}{@{}llll@{}}
\toprule
contig\_a & contig\_b & library\_id & pair\_count \\
\midrule
ctg\_001 & ctg\_187 & mate\_pair & 96 \\
ctg\_001 & ctg\_187 & proximity\_ligation & 192 \\
ctg\_001 & ctg\_187 & short\_insert & 48 \\
\bottomrule
\end{tabular}\par}
\end{tcolorbox}

\begin{tcolorbox}[qaexample,title={Ground-Truth Answer}]
\texttt{ctg\_073}
\end{tcolorbox}
\endgroup

\clearpage
\subsection{Materials Science: Claude Opus 4.8}
\label{app:qa_example_materials_opus48}

\paragraph{Example 1: Specific Surface Area}
The question asks the solver to determine the specific surface area of a porous solid from a nitrogen adsorption isotherm using the multi-point BET method.
The generator simulates a nitrogen adsorption isotherm using the BET model, introduces deviations outside the selected BET fitting region, and stores equilibrium pressures and total adsorbed quantities in \texttt{isotherm.csv}.
It provides the sample mass and physical constants in the question while leaving the valid fitting region to be identified from the data.

\begingroup
\tcbset{qaexample/.style={
    enhanced,
    breakable,
    colback=white,
    colframe=black!35,
    colbacktitle=black!5,
    coltitle=black,
    fonttitle=\small\bfseries,
    fontupper=\small,
    boxrule=0.5pt,
    arc=1mm,
    left=8pt,
    right=8pt,
    top=6pt,
    bottom=6pt,
    before skip=7pt,
    after skip=7pt
}}

\begin{tcolorbox}[qaexample,title={Question},before upper={\raggedright}]
A porous solid was characterized by nitrogen physisorption at 77 K. The file \texttt{isotherm.csv} contains the full measured adsorption isotherm across a wide range of equilibrium pressures. Each row gives the equilibrium absolute pressure (column \texttt{equilibrium\_pressure\_P\_torr}, in Torr) and the corresponding total adsorbed quantity for the entire sample (column \texttt{adsorbed\_quantity\_q\_cm3STP\_total}, in cm3 STP for the whole sample, i.e. NOT normalized per gram). Because the tabulated quantity is the total for the whole sample, the sample mass is required to obtain a per-gram result. Use the following fixed constants: saturation pressure P0 = 98.70 kPa; the exact pressure-unit definition 1 atm = 760 Torr = 101.325 kPa; nitrogen adsorbate cross-sectional area = 0.162 nm2 (0.162e-18 m2 per molecule); STP molar volume = 22414 cm3/mol; Avogadro's number = 6.02214076e23 /mol; sample mass = 0.2500 g; nitrogen molar mass = 28.0134 g/mol. What is the specific surface area, in m2/g, determined by the standard multi-point BET method? Report a single number rounded to one decimal place using round-half-up, with no units and no thousands separator (e.g. '1043.7').
\end{tcolorbox}

\begin{tcolorbox}[qaexample,title={Provided Data (Excerpts)},before upper={\raggedright}]
\textbf{\texttt{isotherm.csv}}\par\smallskip
{\raggedright
\ttfamily\footnotesize
\setlength{\tabcolsep}{5pt}
\renewcommand{\arraystretch}{1.12}
\begin{tabular}{@{}ll@{}}
\toprule
equilibrium\_pressure\_P\_torr & adsorbed\_quantity\_q\_cm3STP\_total \\
\midrule
22.2093 & 13.495 \\
37.0155 & 15.0208 \\
44.4187 & 15.3635 \\
55.5233 & 15.0574 \\
\bottomrule
\end{tabular}\par}
\end{tcolorbox}

\begin{tcolorbox}[qaexample,title={Ground-Truth Answer}]
\texttt{275.0}
\end{tcolorbox}
\endgroup

\clearpage
\paragraph{Example 2: Optical Band Gap}
The question asks the solver to determine the fundamental optical band gap of a semiconductor without a supplied transition type.
The generator simulates an absorption spectrum with an exponential background, a defect-related peak, and multiple interband absorption edges, storing the values in \texttt{absorption\_spectrum.csv}.
It places the fundamental edge on the exponential background and includes a lower-energy defect peak and higher-energy absorption edges to prevent direct identification of the gap from a single visible feature.

\begingroup
\tcbset{qaexample/.style={
    enhanced,
    breakable,
    colback=white,
    colframe=black!35,
    colbacktitle=black!5,
    coltitle=black,
    fonttitle=\small\bfseries,
    fontupper=\small,
    boxrule=0.5pt,
    arc=1mm,
    left=8pt,
    right=8pt,
    top=6pt,
    bottom=6pt,
    before skip=7pt,
    after skip=7pt
}}

\begin{tcolorbox}[qaexample,title={Question},before upper={\raggedright}]
The file \texttt{absorption\_spectrum.csv} contains an optical absorption spectrum measured for a semiconductor thin film. Column 1 is the photon energy in eV (header \texttt{photon\_energy\_eV}) and column 2 is the absorption coefficient in $\mathrm{cm}^{-1}$ (header \texttt{absorption\_coefficient\_cm-1}). The interband transition type of this material is not provided and must be inferred from the data alone. Determine the FUNDAMENTAL optical band gap of the film, defined as the minimum-energy interband transition. Report it in eV as a single number rounded to exactly two decimal places (a plain decimal string, with no units, sign, whitespace, or other text).
\end{tcolorbox}

\begin{tcolorbox}[qaexample,title={Provided Data (Excerpts)},before upper={\raggedright}]
\textbf{\texttt{absorption\_spectrum.csv}}\par\smallskip
{\ttfamily\footnotesize
\setlength{\tabcolsep}{5pt}
\renewcommand{\arraystretch}{1.12}
\begin{tabular}{@{}ll@{}}
\toprule
photon\_energy\_eV & absorption\_coefficient\_cm-1 \\
\midrule
1.20 & 7.48739 \\
1.21 & 8.27485 \\
1.22 & 9.14512 \\
1.23 & 10.1069 \\
1.24 & 11.1699 \\
\bottomrule
\end{tabular}\par}
\end{tcolorbox}

\begin{tcolorbox}[qaexample,title={Ground-Truth Answer}]
\texttt{1.73}
\end{tcolorbox}
\endgroup

\clearpage
\subsection{Materials Science: GPT-5.6-Sol}
\label{app:qa_example_materials_gpt56sol}

\paragraph{Example 1: Synthesis Route Selection}
The question asks the solver to select the precursor route that satisfies the specified conditions under a reaction network inferred from reference observations.
The generator simulates diffraction-channel signals, phase fractions, and heat values from a constructed reaction network and supplies candidate precursor routes and heating programs.
It withholds the active reaction mechanism while providing candidate reactions and simulation rules, so that the reference observations determine how the routes should be evaluated.

\begingroup
\tcbset{qaexample/.style={
    enhanced,
    breakable,
    colback=white,
    colframe=black!35,
    colbacktitle=black!5,
    coltitle=black,
    fonttitle=\small\bfseries,
    fontupper=\small,
    boxrule=0.5pt,
    arc=1mm,
    left=8pt,
    right=8pt,
    top=6pt,
    bottom=6pt,
    before skip=7pt,
    after skip=7pt
}}

\begin{tcolorbox}[qaexample,title={Question},before upper={\raggedright}]
Using \texttt{phase\_catalog.csv}, \texttt{gas\_catalog.csv}, \texttt{visible\_reactions.csv}, \texttt{latent\_slots.json}, \texttt{mechanism\_templates.json}, \texttt{reference\_batches.csv}, \texttt{heating\_schedule.csv}, \texttt{phase\_observation\_matrix.csv}, \texttt{reference\_diffraction.csv}, \texttt{reference\_phase\_fractions.csv}, \texttt{reference\_calorimetry.csv}, \texttt{candidate\_routes.csv}, \texttt{counterfactual\_programs.csv}, and \texttt{rules.json}, identify the sole candidate precursor route satisfying the exact counterfactual qualification rule under the latent reaction network and mechanism uniquely supported by all reference evidence. Return exactly one ASCII route identifier matching \texttt{\textasciicircum{}ROUTE\_[A-R]\$}, with no leading or trailing whitespace and no surrounding text.
\end{tcolorbox}

\begin{tcolorbox}[qaexample,title={Provided Data (Excerpts)},before upper={\raggedright}]
\textbf{\texttt{candidate\_routes.csv}}\par\smallskip
{\raggedright
\ttfamily\footnotesize
\setlength{\tabcolsep}{5pt}
\renewcommand{\arraystretch}{1.12}
\begin{tabular}{@{}lll@{}}
\toprule
route\_id & phase\_id & amount \\
\midrule
ROUTE\_A & P08 & 1 \\
ROUTE\_A & P20 & 1 \\
ROUTE\_A & P26 & 30 \\
\bottomrule
\end{tabular}\par}

\medskip
\textbf{\texttt{reference\_diffraction.csv}}\par\smallskip
{\raggedright
\ttfamily\footnotesize
\setlength{\tabcolsep}{5pt}
\renewcommand{\arraystretch}{1.12}
\begin{tabular}{@{}llll@{}}
\toprule
batch\_id & tick & channel\_id & signal \\
\midrule
BATCH\_1 & 1 & CH1 & 5 \\
BATCH\_1 & 1 & CH2 & 8 \\
BATCH\_1 & 1 & CH3 & 9 \\
\bottomrule
\end{tabular}\par}

\medskip
\textbf{\texttt{reference\_calorimetry.csv}}\par\smallskip
{\raggedright
\ttfamily\footnotesize
\setlength{\tabcolsep}{5pt}
\renewcommand{\arraystretch}{1.12}
\begin{tabular}{@{}lll@{}}
\toprule
batch\_id & tick & signed\_heat \\
\midrule
BATCH\_1 & 1 & 0 \\
BATCH\_1 & 2 & 480 \\
BATCH\_1 & 3 & 0 \\
\bottomrule
\end{tabular}\par}
\end{tcolorbox}

\begin{tcolorbox}[qaexample,title={Ground-Truth Answer}]
\texttt{ROUTE\_M}
\end{tcolorbox}
\endgroup

\clearpage
\paragraph{Example 2: Stress Tensor Analysis}
The question asks the solver to compute the designated stress component in the sensor coordinate frame from material, deformation, and calibration data.
The generator simulates stiffness and deformation data using exact rational values and constructs the associated coordinate transformations and metadata tables.
It distributes the target-case designation, applicable calibration records, and transformation rules across these files so that the stress calculation depends on resolving and combining the relevant inputs.

\begingroup
\tcbset{qaexample/.style={
    enhanced,
    breakable,
    colback=white,
    colframe=black!35,
    colbacktitle=black!5,
    coltitle=black,
    fonttitle=\small\bfseries,
    fontupper=\small,
    boxrule=0.5pt,
    arc=1mm,
    left=8pt,
    right=8pt,
    top=6pt,
    bottom=6pt,
    before skip=7pt,
    after skip=7pt
}}

\begin{tcolorbox}[qaexample,title={Question},before upper={\raggedright}]
Using \texttt{reports.csv}, \texttt{specimens.csv}, \texttt{lots.csv}, \texttt{calibration\_revisions.csv}, \texttt{environment\_states.csv}, \texttt{stiffness\_nodes.csv}, \texttt{deformation\_nodes.csv}, \texttt{frame\_edges.csv}, and \texttt{convention.json} as the authoritative inputs, determine the metadata-designated target case's applicable report-selected one-based sensor-frame Cauchy-stress component. Return exactly one ASCII string matching the regular expression \texttt{\textasciicircum{}sigma\_sensor\_23\_GPa=[+-][0-9]+\textbackslash{}.[0-9]\{12\}\$}, with no spaces and trailing zeros retained. Round the exact rational magnitude to twelve decimal places using round-half-up, then restore the sign.
\end{tcolorbox}

\begin{tcolorbox}[qaexample,title={Provided Data (Excerpts)},before upper={\raggedright}]
\textbf{\texttt{reports.csv}} (selected columns)\par\smallskip
{\ttfamily\footnotesize
\setlength{\tabcolsep}{5pt}
\renewcommand{\arraystretch}{1.12}
\begin{tabular}{@{}llll@{}}
\toprule
report\_id & status & run\_id & sensor\_frame\_id \\
\midrule
R19-v0 & superseded & run\_607 & sensor\_E \\
R19-v1 & accepted & run\_118 & sensor\_A \\
R19-v2 & accepted & run\_431 & sensor\_P \\
R19-v3 & draft & run\_502 & sensor\_D \\
R19-v4 & accepted & run\_304 & sensor\_B \\
\bottomrule
\end{tabular}\par}

\medskip
\textbf{\texttt{stiffness\_nodes.csv}} (selected columns)\par\smallskip
{\ttfamily\footnotesize
\setlength{\tabcolsep}{5pt}
\renewcommand{\arraystretch}{1.12}
\begin{tabular}{@{}llll@{}}
\toprule
calibration\_id & slot\_row & slot\_col & value\_GPa \\
\midrule
cal\_Alpha\_r1 & s0 & s0 & 166 \\
cal\_Alpha\_r1 & s0 & s1 & 27 \\
cal\_Alpha\_r1 & s0 & s2 & -6 \\
cal\_Alpha\_r1 & s0 & s3 & 33 \\
\bottomrule
\end{tabular}\par}

\medskip
\textbf{\texttt{frame\_edges.csv}}\par\smallskip
{\ttfamily\footnotesize
\setlength{\tabcolsep}{4pt}
\renewcommand{\arraystretch}{1.12}
\begin{tabular}{@{}llllll@{}}
\toprule
child\_frame\_id & parent\_frame\_id & i & I & numerator & denominator \\
\midrule
crystal\_118 & fixture\_B & 1 & 1 & 12 & 13 \\
crystal\_118 & fixture\_B & 1 & 2 & -40 & 221 \\
crystal\_118 & fixture\_B & 1 & 3 & 75 & 221 \\
crystal\_118 & fixture\_B & 2 & 1 & 5 & 13 \\
\bottomrule
\end{tabular}\par}
\end{tcolorbox}

\begin{tcolorbox}[qaexample,title={Ground-Truth Answer}]
\texttt{sigma\_sensor\_23\_GPa=-8.737014038398}
\end{tcolorbox}
\endgroup

\clearpage
\subsection{Clinical Imaging: Claude Opus 4.8}
\label{app:qa_example_clinical_opus48}

\paragraph{Example 1: Cardiothoracic Ratio}
The question asks the solver to calculate the cardiothoracic ratio from anatomical landmark measurements associated with a frontal chest radiograph.
The generator reuses a chest radiograph from the MedCTA seed benchmark~\citep{ashraf2026medcta} as \texttt{chest\_pa.png} and constructs \texttt{landmarks.json} with anatomical landmark coordinates and pixel-spacing values.\footnote{Image reused from \citet{ashraf2026medcta}. Sourced from public, de-identified assets under the \href{https://www.apache.org/licenses/LICENSE-2.0}{Apache 2.0} license. Benchmark performance does not indicate clinical readiness.}
It specifies in the question that the ratio should be calculated from the constructed JSON measurements rather than estimated directly from the image.

\begingroup
\tcbset{qaexample/.style={
    enhanced,
    breakable,
    colback=white,
    colframe=black!35,
    colbacktitle=black!5,
    coltitle=black,
    fonttitle=\small\bfseries,
    fontupper=\small,
    boxrule=0.5pt,
    arc=1mm,
    left=8pt,
    right=8pt,
    top=6pt,
    bottom=6pt,
    before skip=7pt,
    after skip=7pt
}}

\begin{tcolorbox}[qaexample,title={Question},before upper={\raggedright}]
You are given a frontal (PA) chest radiograph, \texttt{chest\_pa.png}, together with a machine-readable annotation file, \texttt{landmarks.json}. The JSON file provides, in image pixel coordinates ([x, y], origin top-left), multiple sampled candidate points along each of several labeled anatomical structures on this radiograph, plus several additional individually labeled annotations and the image pixel-spacing values (in millimeters) for the x and y axes. Based solely on the provided measurements in \texttt{landmarks.json}, compute this patient's cardiothoracic ratio and report its exact value. Answer with a single line containing only the ratio rounded to exactly three decimal places, written with a single leading zero and exactly three digits after the decimal point (for example, 0.543). No percent sign, label, or other text.
\end{tcolorbox}

\begin{tcolorbox}[qaexample,title={Provided Data (Excerpts)},before upper={\raggedright}]
\noindent\begin{minipage}[t]{0.45\linewidth}
\vspace{0pt}
\textbf{\texttt{chest\_pa.png}}\par\smallskip
\noindent\includegraphics[width=1\linewidth]{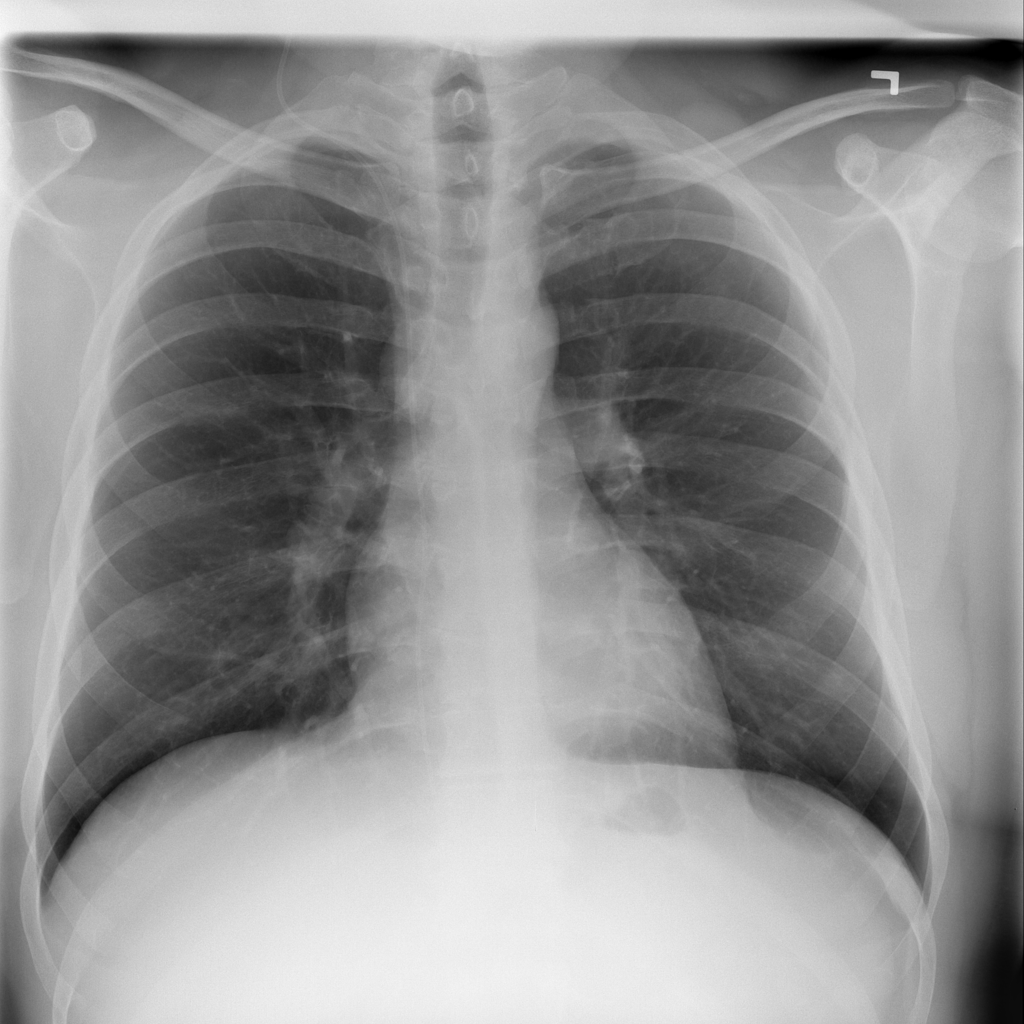}\par
\end{minipage}\hfill
\begin{minipage}[t]{0.52\linewidth}
\vspace{0pt}
\textbf{\texttt{landmarks.json}}\par\smallskip
\begin{lstlisting}[language={},basicstyle=\ttfamily\footnotesize,columns=fullflexible,keepspaces=true,showstringspaces=false,breaklines=true,aboveskip=0pt,belowskip=0pt]
{
  "right_cardiac_border": [
    [350, 640], [340, 558],
    [347, 598], ...
  ],
  "left_cardiac_border": [
    [700, 500], [712, 566],
    [706, 540], ...
  ],
  "pixel_spacing_x_mm": 0.2,
  "pixel_spacing_y_mm": 0.2
}
\end{lstlisting}
\end{minipage}\par
\end{tcolorbox}

\begin{tcolorbox}[qaexample,title={Ground-Truth Answer}]
\texttt{0.540}
\end{tcolorbox}
\endgroup

\clearpage
\paragraph{Example 2: Treatment Response}
The question asks the solver to determine the RECIST 1.1 response category at the latest follow-up from lesion annotations across three time points.
The generator reuses chest and abdominal CT images from MedCTA~\citep{ashraf2026medcta} and adds constructed lesion identities, measurements, and dates to create a synthetic longitudinal case, rather than using an actual patient follow-up series.\footnote{Images adapted from \citet{ashraf2026medcta}. Sourced from public, de-identified assets under the \href{https://www.apache.org/licenses/LICENSE-2.0}{Apache 2.0} license. Benchmark performance does not indicate clinical readiness.}
It places the lesion annotations directly on the images and preserves lesion identifiers across time points so that changes in lesion size and the appearance of new lesions can be compared.

\begingroup
\tcbset{qaexample/.style={
    enhanced,
    breakable,
    colback=white,
    colframe=black!35,
    colbacktitle=black!5,
    coltitle=black,
    fonttitle=\small\bfseries,
    fontupper=\small,
    boxrule=0.5pt,
    arc=1mm,
    left=8pt,
    right=8pt,
    top=6pt,
    bottom=6pt,
    before skip=7pt,
    after skip=7pt
}}

\begin{tcolorbox}[qaexample,title={Question},before upper={\raggedright}]
You are given three axial CT studies from a single oncology patient, acquired at three timepoints: a baseline study and two follow-up studies (files: \texttt{baseline\_ct.png}, \texttt{followup1\_ct.png}, \texttt{followup2\_ct.png}). Each image is a dated two-panel composite (a chest panel and an abdomen panel) showing the patient's lesions on real CT anatomy. Every lesion identified at a given timepoint is annotated inline in that study with a caliper and a text label giving the lesion ID, its organ/location, its lesion type (parenchymal lesion or lymph node), and its measured dimensions in millimeters \textemdash{} parenchymal lesions are labeled with a long-axis diameter, and lymph nodes are labeled with both a long-axis and a short-axis diameter. Each lesion keeps the same ID across the studies in which it appears, so a study is annotated with exactly the lesions present at that timepoint (the absence of an ID from a study means that lesion was not present then). Using RECIST 1.1, determine the overall response category at Follow-up 2 (the latest study, \texttt{followup2\_ct.png}). Answer with exactly one of: CR, PR, SD, PD.
\end{tcolorbox}

\begin{tcolorbox}[qaexample,title={Provided Data},before upper={\raggedright}]
\noindent\begin{minipage}[t]{0.32\linewidth}
\vspace{0pt}
\textbf{\texttt{baseline\_ct.png}}\par\smallskip
\includegraphics[width=\linewidth]{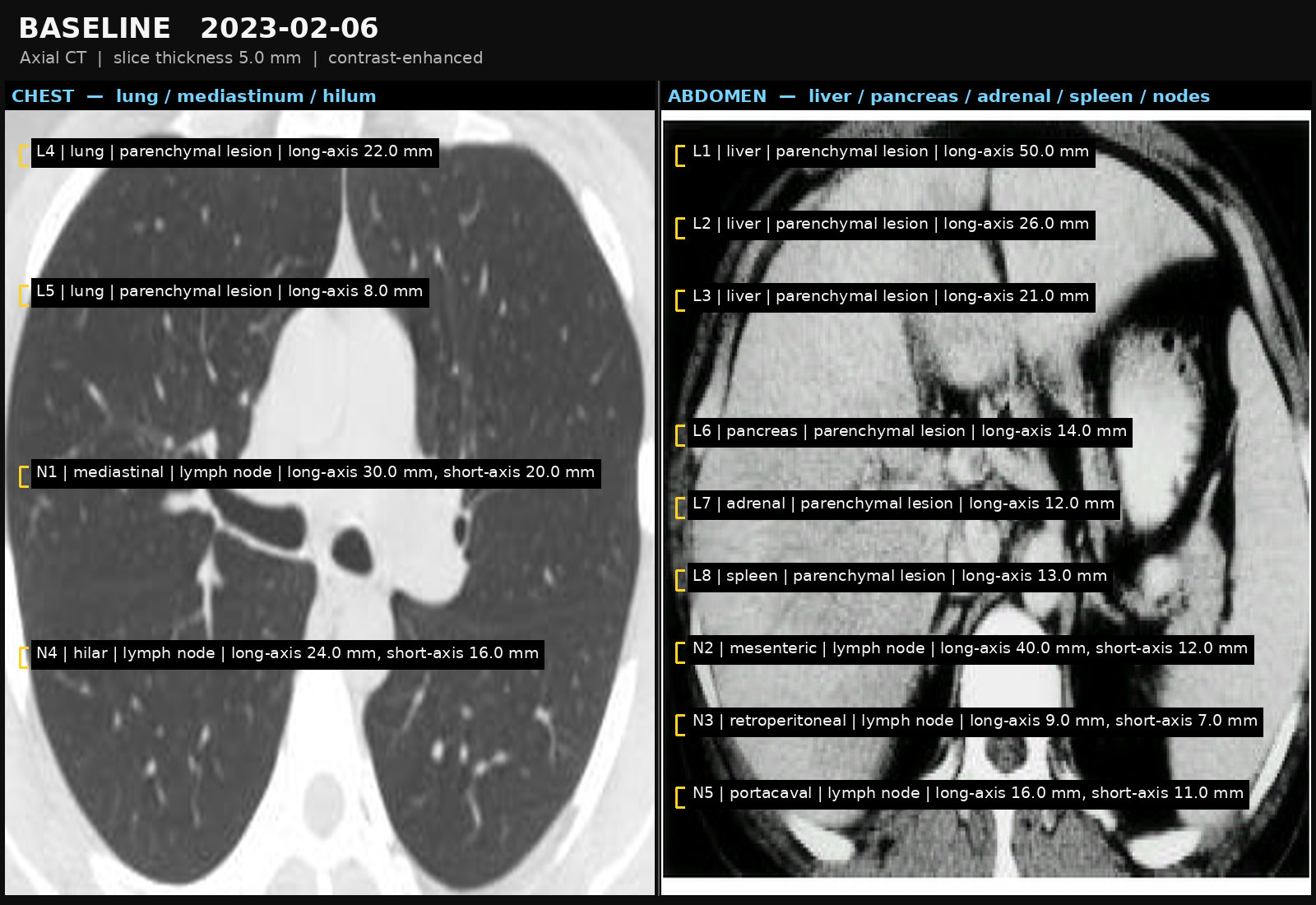}
\end{minipage}\hfill
\begin{minipage}[t]{0.32\linewidth}
\vspace{0pt}
\textbf{\texttt{followup1\_ct.png}}\par\smallskip
\includegraphics[width=\linewidth]{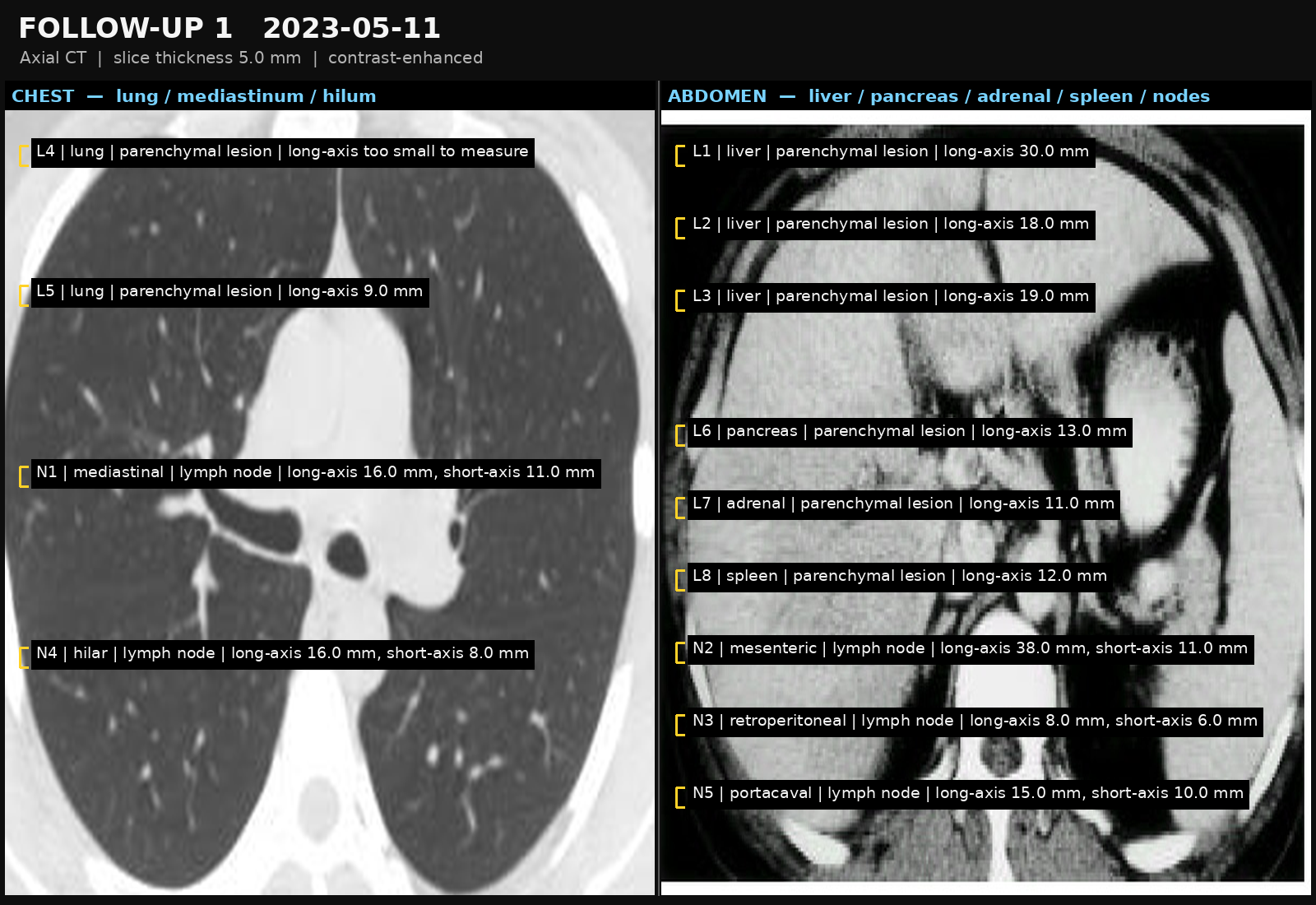}
\end{minipage}\hfill
\begin{minipage}[t]{0.32\linewidth}
\vspace{0pt}
\textbf{\texttt{followup2\_ct.png}}\par\smallskip
\includegraphics[width=\linewidth]{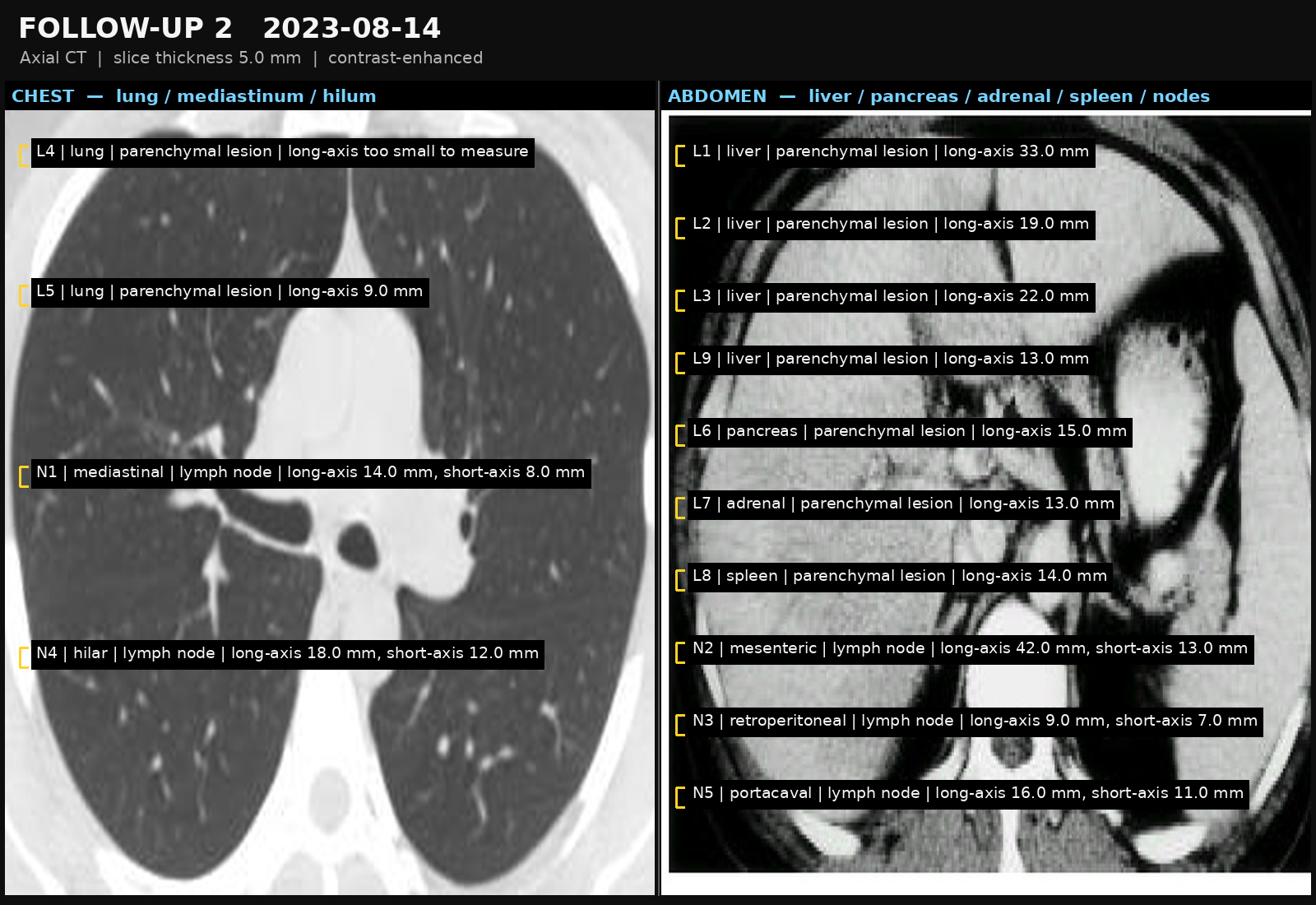}
\end{minipage}\par
\end{tcolorbox}

\begin{tcolorbox}[qaexample,title={Ground-Truth Answer}]
\texttt{PD}
\end{tcolorbox}
\endgroup

\clearpage
\subsection{Clinical Imaging: GPT-5.6-Sol}
\label{app:qa_example_clinical_gpt56sol}

\paragraph{Example 1: Cardiac MRI Interpretation}
The question asks the solver to classify the spatial correspondence and coronary distribution of edema and enhancement in cardiac MRI.
The generator fetches a cardiac MRI montage from an open-access publication on the web~\citep{gulhane2023cardiac} and provides it as \texttt{case\_cmr\_montage.png}.\footnote{Image adapted from \citet{gulhane2023cardiac}. Sourced from public, de-identified assets under the \href{https://creativecommons.org/licenses/by/4.0/}{CC BY 4.0} license. Benchmark performance does not indicate clinical readiness.}
It constructs \texttt{study\_metadata.json} to identify the imaging sequences, views, and display orientation, and \texttt{aha16\_orientation\_legend.png} to provide an anatomical segment reference, while leaving the spatial patterns to be interpreted from the montage.

\begingroup
\tcbset{qaexample/.style={
    enhanced,
    breakable,
    colback=white,
    colframe=black!35,
    colbacktitle=black!5,
    coltitle=black,
    fonttitle=\small\bfseries,
    fontupper=\small,
    boxrule=0.5pt,
    arc=1mm,
    left=8pt,
    right=8pt,
    top=6pt,
    bottom=6pt,
    before skip=7pt,
    after skip=7pt
}}

\begin{tcolorbox}[qaexample,title={Question},before upper={\raggedright}]
Given \texttt{case\_cmr\_montage.png}, \texttt{aha16\_orientation\_legend.png}, and \texttt{study\_metadata.json}, what is the combined edema--enhancement spatial-concordance and coronary-distribution class for this study? Return exactly one lowercase ASCII string from: concordant\_coronary, concordant\_noncoronary, discordant\_coronary, or discordant\_noncoronary.
\end{tcolorbox}

\begin{tcolorbox}[qaexample,title={Provided Data (Excerpts)},before upper={\raggedright}]
\textbf{\texttt{case\_cmr\_montage.png}}\par\smallskip
\noindent\includegraphics[width=0.67\linewidth]{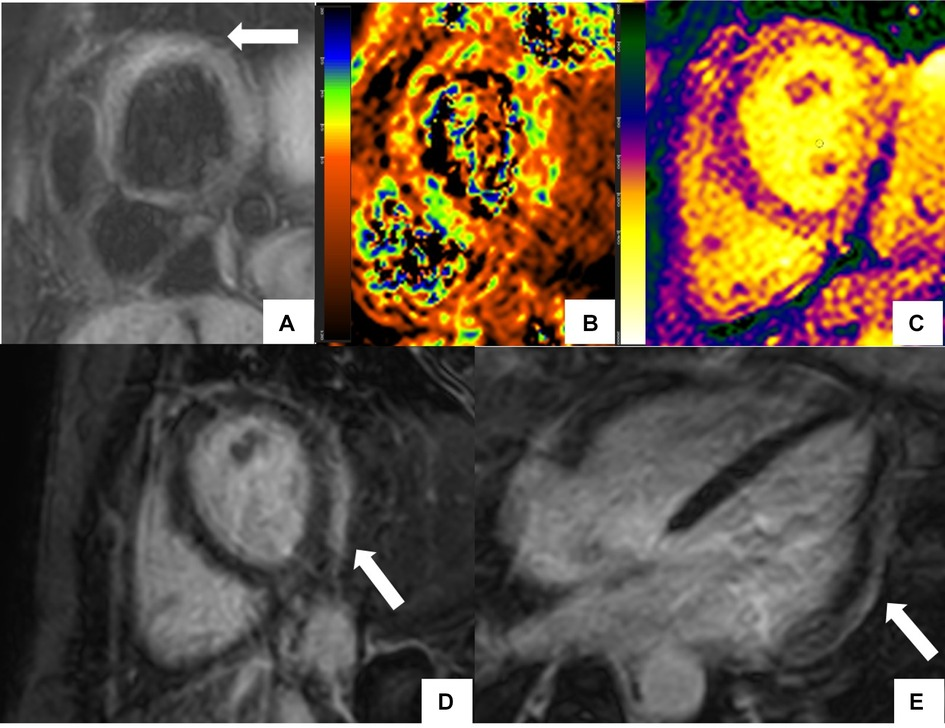}\par

\medskip
\textbf{\texttt{study\_metadata.json}}\par\smallskip
\begin{lstlisting}[language={},basicstyle=\ttfamily\footnotesize,columns=fullflexible,keepspaces=true,showstringspaces=false,breaklines=true,aboveskip=0pt,belowskip=0pt]
{
  "same_examination": true,
  "short_axis_display": {
    "anterior": "top",
    "inferior": "bottom",
    "patient_right": "image_left",
    "patient_left": "image_right"
  },
  "aha_model": "16-segment model; apex excluded"
}
\end{lstlisting}
\end{tcolorbox}

\begin{tcolorbox}[qaexample,title={Ground-Truth Answer}]
\texttt{concordant\_noncoronary}
\end{tcolorbox}
\endgroup

\clearpage
\paragraph{Example 2: Spatial Cell Composition}
The question asks the solver to determine the cell-type composition of a selected anchor cell's spatial neighborhood from multiplex immunofluorescence images.
The generator reuses eight marker channels and the aligned cell-instance mask from the CellSighter test dataset, derived from the CRC-FFPE-CODEX collection~\citep{bussi2025cellsighter,schurch2020crccodex}.\footnote{Images and masks adapted from \citet{bussi2025cellsighter,schurch2020crccodex}. Sourced from public, de-identified assets under the \href{https://creativecommons.org/licenses/by/4.0/}{CC BY 4.0} license. Benchmark performance does not indicate clinical readiness.}
It constructs \texttt{definitions.json} with marker-positivity criteria, phenotype-assignment rules, and spatial-neighborhood definitions, while leaving cell-level marker signals and spatial relationships to be recovered from the images and mask.

\begingroup
\tcbset{qaexample/.style={
    enhanced,
    breakable,
    colback=white,
    colframe=black!35,
    colbacktitle=black!5,
    coltitle=black,
    fonttitle=\small\bfseries,
    fontupper=\small,
    boxrule=0.5pt,
    arc=1mm,
    left=8pt,
    right=8pt,
    top=4pt,
    bottom=4pt,
    before skip=4pt,
    after skip=4pt
}}

\begin{tcolorbox}[qaexample,title={Question},before upper={\raggedright}]
Using \texttt{marker\_channels.ome.tif}, \texttt{cell\_labels.tif}, and \texttt{definitions.json}, what is the phenotype composition of the definitions-selected anchor cell's defined spatial neighborhood? Return exactly one ASCII string formatted as \texttt{tumor=\textless{}integer\textgreater{};\allowbreak CD4\_T=\textless{}integer\textgreater{};\allowbreak CD8\_T=\textless{}integer\textgreater{};\allowbreak Treg=\textless{}integer\textgreater{};\allowbreak B\_cell=\textless{}integer\textgreater{};\allowbreak macrophage=\textless{}integer\textgreater{};\allowbreak dendritic=\textless{}integer\textgreater{};\allowbreak other=\textless{}integer\textgreater{}}.
\end{tcolorbox}

\begin{tcolorbox}[qaexample,title={Provided Data (Excerpts)},before upper={\raggedright}]
\noindent\begin{minipage}[t]{0.48\linewidth}
\textbf{\texttt{marker\_channels.ome.tif}}\par\smallskip
\includegraphics[width=\linewidth]{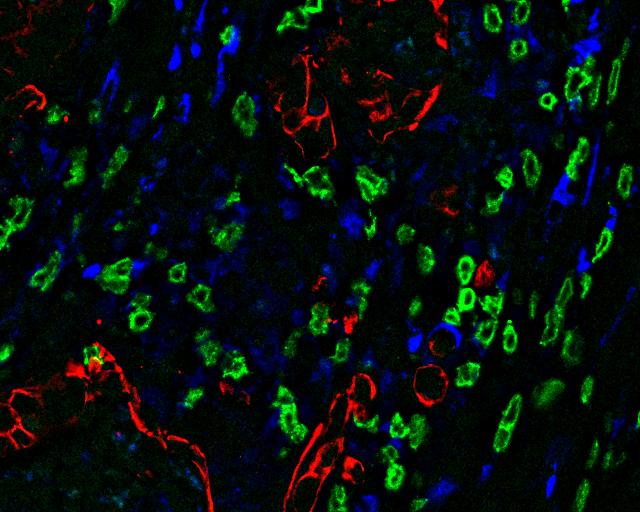}
\end{minipage}\hfill
\begin{minipage}[t]{0.48\linewidth}
\textbf{\texttt{cell\_labels.tif}}\par\smallskip
\includegraphics[width=\linewidth]{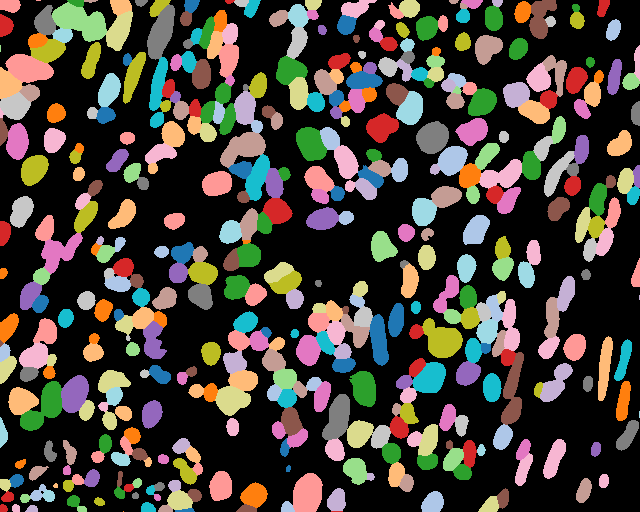}
\end{minipage}\par
\smallskip
{\footnotesize The same region is shown in both panels. Left: Cytokeratin (red), CD3 (green), and CD68 (blue), independently contrast-scaled for display. Right: cell-instance labels shown in pseudocolor; colors do not indicate phenotypes.}\par

\medskip
\textbf{\texttt{definitions.json}} (selected fields)\par\smallskip
\begin{lstlisting}[language={},basicstyle=\ttfamily\footnotesize,columns=fullflexible,keepspaces=true,showstringspaces=false,breaklines=true,aboveskip=0pt,belowskip=0pt]
{
  "marker_calling": {
    "positive_rule": "A retained label is marker-positive exactly when 2*s2 >= p2+n2 for that marker."
  },
  "phenotype_assignment": {
    "semantics": "Evaluate rules in ascending precedence and assign exactly the first matching phenotype."
  },
  "answer_neighborhood": {
    "membership": "The union of retained vertices at shortest unweighted graph-path distance exactly one or exactly two from the selected anchor. Exclude the anchor and count each label once."
  }
}
\end{lstlisting}
\end{tcolorbox}

\begin{tcolorbox}[qaexample,title={Ground-Truth Answer},before upper={\raggedright}]
\texttt{tumor=2;\allowbreak CD4\_T=3;\allowbreak CD8\_T=9;\allowbreak Treg=8;\allowbreak B\_cell=1;\allowbreak macrophage=19;\allowbreak dendritic=17;\allowbreak other=47}
\end{tcolorbox}
\endgroup

\clearpage
\section{Prompts}
\label{app:agent_prompts}

This section lists the prompts for concept generation, task instantiation, solving and judging, refinement, experience-guided generation, and benchmark quality assessment. Instance-specific inputs are represented by placeholders.

\subsection{Initial Concept Generation}
\label{app:prompt_initial_concepts}

\paragraph{Benchmark Concept Extraction}
Fig.~\ref{fig:prompt_benchmark_concept_extraction} shows the prompt for extracting a concept from a seed question and its associated file information.

\paragraph{Literature Search and Workflow Extraction}
Fig.~\ref{fig:prompt_workflow_discovery} shows the prompt for searching scientific literature and extracting analytical workflows.

\paragraph{Pattern Extraction}
Fig.~\ref{fig:prompt_pattern_extraction} shows the prompt for identifying recurring analytical patterns from the extracted workflows.

\paragraph{Concept Generation}
Fig.~\ref{fig:prompt_initial_concept_generation} shows the prompt for generating concepts from benchmark-derived concepts, workflows, and patterns.

\subsection{Task Instantiation}
\label{app:prompt_instantiation}

\paragraph{Recipe Planning}
Fig.~\ref{fig:prompt_recipe_planning} shows the prompt for translating a concept into a recipe for task construction.

\paragraph{Task Implementation}
Fig.~\ref{fig:prompt_qa_implementation} shows the prompt for constructing a question, associated data, and reference answer from a concept and recipe.

\subsection{Multi-Trial Solving and Judging}
\label{app:prompt_solving_judging}

\paragraph{Solving}
Fig.~\ref{fig:prompt_solving} shows the prompt for solving a task given its question, associated files, and answer format.

\paragraph{Judging}
Fig.~\ref{fig:prompt_judging} shows the prompt for assessing task verifiability and difficulty using the reference answer and solver results and traces.

\subsection{Feedback-Guided Refinement}
\label{app:prompt_refinement}

\paragraph{Recipe Revision}
Fig.~\ref{fig:prompt_recipe_refinement} shows the prompt for revising a recipe using the implemented task and evaluation or implementation feedback.

\paragraph{Concept Revision}
Fig.~\ref{fig:prompt_concept_refinement} shows the prompt for revising a concept based on recent refinement history.

\paragraph{Implementation after Refinement}
Fig.~\ref{fig:prompt_refined_qa_implementation} shows the prompt for implementing the revised recipe using the previous task, available data, and judge feedback.

\subsection{Experience-Guided Concept Generation}
\label{app:prompt_experience}

\paragraph{Experience Construction}
Fig.~\ref{fig:prompt_experience_construction} shows the prompt for distilling reusable design guidance from completed refinement histories.

\paragraph{Generation with Experience}
Fig.~\ref{fig:prompt_experience_guidance} shows the experience-guidance block used with the concept-generation prompt in Fig.~\ref{fig:prompt_initial_concept_generation}.

\subsection{Benchmark Quality Assessment}
\label{app:prompt_quality_assessment}

Fig.~\ref{fig:prompt_quality_assessment} shows the prompt for assessing verifiability, usefulness, and fitness from a question and its associated data files.

\subsection{Domain-Specific Prompt Variants}
\label{app:prompt_domain_variants}

Figures~\ref{fig:prompt_computational_biology_substitutions}, \ref{fig:prompt_materials_science_substitutions}, and~\ref{fig:prompt_medical_imaging_substitutions} provide domain-specific settings for the shared prompts. Fig.~\ref{fig:prompt_medical_imaging_planning_guidance} shows the shared guidance for clinical-imaging recipe planning and revision.

\begingroup
\captionsetup{type=figure,position=bottom,skip=4pt}
\begin{promptbox}
\begin{lstlisting}[style=agentprompttext]
Label this benchmark question with its concept = (domain, modality, operation).

Definitions:
- domain: the {field} the question is about (e.g. {domain_examples}).
- modality: the data form the task operates on (e.g. {modality_examples}, or "knowledge" if there is no input file). Use a short lowercase token.
- operation: THE ATOMIC KNOWLEDGE / CAPABILITY / REASONING TARGET THE QUESTION IS INTENDED TO TEST — the single unit the QA is probing, named as a noun phrase at the most abstract level. Capture the testable IDEA, not what an agent does and not how it solves it.

Output CONTRACT — return ONLY one JSON object, no prose, no markdown fences:
{"domain": "<short>", "modality": "<short token>", "operation": ["<target 1>", "<target 2>", ...], "rationale": ...}

QUESTION_ID: {question_id}
FILE_PATHS: {file_paths}
QUESTION: {question}

You are an expert curator of a {benchmark_subject} agent benchmark for the {field}. You label each benchmark question with its underlying CONCEPT.
\end{lstlisting}
\end{promptbox}
\nopagebreak
\caption{Prompt for benchmark concept extraction.}
\label{fig:prompt_benchmark_concept_extraction}
\par
\endgroup

\begingroup
\captionsetup{type=figure,position=bottom,skip=4pt}
\begin{promptbox}
\begin{lstlisting}[style=agentprompttext]
You are an autonomous external-source discovery agent for {benchmark_subject} benchmark concept generation. Your primary job is to find source-grounded workflows or experimental flows from external knowledge. Use recent literature as the default evidence source: papers or preprints from {recent_source_cutoff_year} or later. Extract how each study or source actually constructed the workflow from inputs to outputs, conclusions, or estimands. Use standards, databases, benchmark descriptions, and documentation only when they clarify or define a durable workflow, not as substitutes for recent workflow evidence.

DISCOVERY SCOPE
{scope}

ALREADY COLLECTED SOURCES TO AVOID
The following source papers/records have already been collected by earlier calls. Do not return workflows from these exact sources, near-duplicate versions, or obvious re-summarizations of them.
1. title: {source_title} | doi: {source_doi} | pmid: {source_pmid} | url: {source_url}

Task:
Find public external sources that can inspire benchmark concepts by revealing how {benchmark_subject} workflows are actually designed. For each useful source, extract the workflow or experimental flow by which the source reaches a conclusion, result, output, or estimand.

Recency requirement:
- Default source window: papers/preprints published in {recent_source_cutoff_year} or later (year-level approximation of the last five years).
- First search for recent source papers/preprints. Prefer returning fewer workflows or [] over filling the output with older, weakly justified sources.
- Target nearly all returned workflows from {recent_source_cutoff_year} or later.
- Use a source before {recent_source_cutoff_year} only as an exception when it is necessary context for a recent workflow, no recent source exposes the workflow detail, or the source is a durable standard/database/benchmark specification.
- If you use an older source, explicitly explain the exception in `discovery_trace.source_selection_reason`.

Important:
- Return only workflows that are genuinely supported by specific external sources.
- It is acceptable to return a small number of workflows or [] if you cannot find enough well-supported workflow evidence.
- Avoid shallow, duplicate, speculative, or quota-filling records.
- Search for sources not listed in the avoid section above. If the best source you find is already listed there, omit it and look for a different source instead.
- This must be external-knowledge based. Ground every workflow in a specific public source.
- If a source is interesting but does not expose enough workflow/experimental-flow detail, omit it.
- Extract how the workflow was built: input data, preprocessing/experimental setup, inference or analysis steps, intermediate checks, output/estimand, and interpretation.
- Do not copy a benchmark task from a single paper. We are collecting workflow evidence.
- Cite source identifiers when possible: URL, DOI, PMID, accession, standard name, or docs URL.
- If you are uncertain about a source, include `source_confidence`: "low" and explain why.
- If the source uses specific tools, software, packages, pipelines, databases, or assays, record them in `tool_usage`. Do not force tool names into the workflow text when they are incidental, but do preserve them when they are important for understanding the source workflow.
- Include a concise `discovery_trace` explaining how the source/workflow was identified and what evidence in the source supports the extracted workflow.

Output CONTRACT - return ONLY a JSON array, no prose, no markdown fences:
[
  {
    "source": {
      "title": "...",
      "url": "...",
      "doi": "...",
      "pmid": "...",
      "year": "...",
      "source_type": "<paper | review | standard | database | tool_doc | benchmark | other>",
      "source_confidence": "<high | medium | low>"
    },
    "domain": "...",
    "data_modality": ["..."],
    "scientific_goal": "...",
    "workflow": [
      "input/data context",
      "analysis or experimental step",
      "intermediate decision/check",
      "final output/conclusion"
    ],
    "tool_usage": [
      {
        "tool_name": "...",
        "tool_type": "<software | database | assay | benchmark | standard | pipeline | other>",
        "role_in_workflow": "how this tool/source was used in the paper's workflow"
      }
    ],
    "target_estimand": "free-form description of the target quantity, output, conclusion, or estimand, if the source defines one",
    "inputs_required": ["..."],
    "outputs_produced": ["..."],
    "decision_points": ["..."],
    "plausible_wrong_paths": ["..."],
    "conclusion_or_output": "...",
    "discovery_trace": {
      "search_focus": "...",
      "source_selection_reason": "...",
      "evidence_used": ["paraphrased evidence cue", "..."],
      "workflow_extraction_reason": "...",
      "uncertainty": "..."
    },
    "rationale": "why this source/workflow is useful for benchmark concept generation"
  }
]
\end{lstlisting}
\end{promptbox}
\nopagebreak
\caption{Prompt for literature search and workflow extraction.}
\label{fig:prompt_workflow_discovery}
\par
\endgroup

\begingroup
\captionsetup{type=figure,position=bottom,skip=4pt}
\begin{promptbox}
\begin{lstlisting}[style=agentprompttext]
You consolidate source-grounded workflows into recurring high-level analysis or experimental-flow patterns. Patterns are abstract structures across workflows, not specific papers, tools, or implementations.

DISCOVERY SCOPE
{scope}

SOURCE-GROUNDED WORKFLOW CORE RECORDS
Read the workflow core records from this local JSON file before producing patterns:
{workflow_core_records_path}

The JSON object contains `workflow_core_records`, a list of workflow records with:
- workflow_id
- domain
- data_modality
- scientific_goal
- workflow

Task:
Freely extract recurring high-level workflow/experimental-flow patterns from the workflows.

Important:
- Choose the number of patterns yourself. Do not force a fixed count.
- Return [] if the workflow set does not support useful recurring patterns.
- A pattern should abstract across one or more workflows.
- Preserve workflow_ids supporting each pattern.
- Use the full supplied workflow_id range, not only early examples.
- In `workflow_ids`, include all workflows that clearly support the pattern.
- Keep patterns tool-agnostic.

Output CONTRACT — return ONLY a JSON array, no prose, no markdown fences:
[
  {
    "pattern_id": "P1",
    "pattern_label": "...",
    "pattern_definition": "...",
    "workflow_ids": ["W1", "W2"],
    "canonical_estimands": ["..."],
    "recurring_flow": ["..."],
    "recurring_outputs": ["..."],
    "recurring_decision_points": ["..."],
    "plausible_wrong_path_families": ["..."],
    "rationale": "..."
  }
]
\end{lstlisting}
\end{promptbox}
\nopagebreak
\caption{Prompt for pattern extraction.}
\label{fig:prompt_pattern_extraction}
\par
\endgroup

\begingroup
\captionsetup{type=figure,position=bottom,skip=4pt}
\begin{promptbox}
\begin{lstlisting}[style=agentprompttext]
You design concepts for a {benchmark_subject} agent benchmark. A concept is (domain, modality, operation).

You are generating concepts, not selecting from a pre-existing candidate pool. The generated concepts will be sent directly into the QA refinement loop without an additional selection/filtering stage.

EXPERIENCE GUIDANCE
No previous experience cards are provided. Generate the initial small concept batch from the source priors.

SOURCE PRIORS
BENCHMARK-INFERRED CONCEPT PRIOR
{benchmark_concepts_json}

EXTERNAL WORKFLOW PRIOR
{workflow_cores_json}

EXTERNAL PATTERN PRIOR
{patterns_json}

ALREADY USED OR ALREADY PROPOSED CONCEPTS TO AVOID
{previous_concepts_json}

TASK
Propose {n} NEW concepts that can plausibly become benchmark QAs that are both harder and verifiable. Avoid exact duplicates and trivial variants of the avoid list.

Definition:
- `domain`: the {benchmark_subject} area.
- `modality`: the input/data type the future QA will expose.
- `operation`: the atomic knowledge/capability/reasoning target the QA tests. It should be a noun phrase or a compact list of noun phrases naming WHAT is tested, not a step-by-step procedure and not a solver instruction.
- `rationale`: why this concept is promising for a harder and verifiable QA.

Quality requirements:
- The concept should support a single, unambiguous answer after recipe and QA implementation.
- Avoid concepts whose answer would naturally require ranges, tolerances, subjective interpretation, stochastic model fitting, or arbitrary parameter choices unless the future data can force one exact choice.
- Keep operations specific enough to be testable; avoid vague labels such as
  "integrative analysis" without a concrete target.

Output CONTRACT - return ONLY a JSON array of {n} objects, no prose, no fences:
[{"domain": "...", "modality": "...", "operation": ["<target 1>", "<target 2>", "..."], "rationale": "..."}]
\end{lstlisting}
\end{promptbox}
\nopagebreak
\caption{Prompt for initial concept generation.}
\label{fig:prompt_initial_concept_generation}
\par
\endgroup

\begingroup
\captionsetup{type=figure,position=bottom,skip=4pt}
\begin{promptbox}
\begin{lstlisting}[style=agentprompttext]
You are PLANNING how to BUILD ONE benchmark QA for a {benchmark_subject} agent benchmark, from the CONCEPT below. A concept = (domain, modality, operation).

CONCEPT
  domain: {domain}
  modality: {modality}
  operation (atomic target to test): {ops}

Produce a GENERATION RECIPE: the BUILDER's strategy for turning this concept into a real QA with a single, checkable ground-truth answer. This is the how-to-BUILD, NOT the how-to-SOLVE: describe how YOU (the builder) will source or construct the data, make and hide the correct answer, and compute & verify the ground truth. Do NOT write the solver's solution steps — the solver must rediscover the method; you are only planning construction.

Guidance:
- You may REUSE existing benchmark data (a copy is available at: {existing_data}), FETCH new data from the web, or SIMULATE data — choose what best fits the concept and say which.
{medical_imaging_plan_guidance}- The construction mechanism is how the single correct answer comes to exist and stay hidden (an intrinsic property of real data, a planted/simulated signal, a removed label, a permuted label, ...). Be concrete.
- The ground-truth plan must let the builder COMPUTE the answer for real and VERIFY it (recompute, cross-check, or — if simulated — read it off the known generative parameters).
- Exact-match grading is the only grading mode. The answer must be one single string, not a range, tolerance interval, confidence interval, ranked set, distribution, or approximate value. If the answer is numeric, specify the exact rounding/precision rule in answer_spec.
- The final answer and the required intermediate choices must be unique. If solving requires choosing hyperparameters, thresholds, model settings, normalization rules, tie breakers, reference versions, or filtering criteria, the data files must contain clear evidence that forces exactly those choices. If multiple reasonable choices remain, the recipe is invalid.
- Avoid variance. Do not make correctness depend on stochastic model training, random sampling, nondeterministic clustering, or unstable web/database state unless the data construction fixes the seed/version and the ground-truth plan verifies the deterministic result.
- Do not make the benchmark hinge on malicious data-style tricks such as incidental whitespace, lowercase/uppercase differences, row ordering, duplicate labels, or delimiter ambiguity unless that style issue is explicitly the concept being tested. Use canonical formatting, explicit delimiters, stable sorting, and documented normalization where needed.
- The question-hiding strategy should say how the final question will avoid revealing intermediate reasoning, solution steps, or procedural hints, while still specifying enough input files, target quantity, and answer format for a unique answer.
- This is a PLAN ONLY: do not fetch data, run tools, or build anything now.

Output ONLY one JSON object (no prose, no markdown fences):
{"data_strategy": "<reuse existing benchmark data | fetch from web | simulate>",
 "data_plan": "<what data to obtain/build and how it carries the answer>",
 "construction_mechanism": "<how the single correct answer is made & hidden>",
 "ground_truth_plan": "<how to COMPUTE and VERIFY the answer for real, including why intermediate choices are uniquely determined>",
 "answer_spec": "<the answer type + exact string format, for exact-match grading; no ranges/tolerances>",
 "tools_resources": ["<tool / DB / internet / GPU the build needs>", ...],
 "question_hiding_strategy": "<how to phrase the solver-facing question without exposing intermediate reasoning or solution steps, while still giving enough information for a unique answer>"}
\end{lstlisting}
\end{promptbox}
\nopagebreak
\caption{Prompt for recipe planning.}
\label{fig:prompt_recipe_planning}
\par
\endgroup

\begingroup
\captionsetup{type=figure,position=bottom,skip=4pt}
\begin{promptbox}
\begin{lstlisting}[style=agentprompttext]
You are building ONE benchmark question for a {benchmark_subject} agent benchmark, from the CONCEPT below. A concept = (domain, modality, operation).

CONCEPT
  domain: {domain}
  modality: {modality}
  operation (atomic target to test): {ops}

IMPORTANT — `operation` names the ATOMIC TARGET (a knowledge/capability/reasoning target) this question must test, NOT a recipe. YOU must devise a concrete question whose answer genuinely hinges on that target, work out a real solution method, and compute/verify the ground-truth answer yourself. Do NOT reveal your method to the solver — the question must test whether a solver can figure it out.

GENERATION RECIPE — a pre-planned strategy for BUILDING this QA. FOLLOW IT to construct the data and compute/verify the ground truth (adjust implementation details as needed; it is a plan, not the solver's solution). If the recipe is underspecified in a way that would allow multiple answers, tighten the data construction so the answer and intermediate choices become uniquely determined. Do NOT reveal any of it to the solver.
  data strategy: {data_strategy}
  data plan: {data_plan}
  construction mechanism: {construction_mechanism}
  ground-truth plan: {ground_truth_plan}
  answer spec: {answer_spec}
  tools/resources: {tools_resources}
  question hiding strategy: {question_hiding_strategy}


YOUR TASK — actually build a real, answerable question with a KNOWN, CHECKABLE ground-truth answer:
- You are in a writable working directory. You may run code and tools.
- You are free to modify the current conda environment as needed.
- You may REUSE existing benchmark data (a copy is available at: {existing_data}) OR fetch/simulate new data from the web — whichever fits the concept. Possibilities are open; do not restrict yourself.
- If the question needs input data files, create them in THIS working directory.
- Determine the ground-truth answer for real (compute it, look it up, or — if you simulate the data — use the known generative parameters). Do not invent an answer you have not verified.
- Exact-match grading is the only grading mode. The answer must be one single string, not a range, tolerance interval, confidence interval, ranked set, distribution, or approximate value. If the answer is numeric, define and use one exact rounding/precision rule.
- The final answer and the necessary intermediate choices must be unique. If the solution requires choosing hyperparameters, thresholds, model settings, normalization rules, tie breakers, reference versions, or filtering criteria, the solver-facing data files must contain clear evidence that forces exactly those choices. Do not rely on unstated defaults or conventions.
- Avoid variance. Do not make the answer depend on stochastic model training, random sampling, nondeterministic clustering, or changing web/database state unless you fix the seed/version and verify the deterministic answer.
- Do not make correctness hinge on malicious data-style tricks such as incidental whitespace, lowercase/uppercase differences, row ordering, duplicate labels, or delimiter ambiguity unless that style issue is explicitly the concept being tested. Use canonical formatting, explicit delimiters, stable sorting, and documented normalization where needed.

WRITING THE QUESTION TEXT — state only WHAT is being asked, never HOW to solve it:
- Give the solver only the goal, the input file(s), and the required answer format.
- DO NOT list, number, or paraphrase the solution method; do NOT name the specific tools, commands, models, algorithms, or parameters you used; do NOT outline a procedure or pipeline. No "first… then…" walkthroughs, no step lists, no hints at the method.
- A good question reads like a real research request ("Given <file>, what is <quantity>?"), leaving the approach entirely to the solver.
- Every solver-facing input file must be named explicitly in the question text. Put only those explicitly named solver-facing input files in `data_files`; do not include build scripts, verification scripts, answer keys, notes, logs, scratch files, intermediate outputs, or helper/index files unless the question explicitly names them as inputs.

When done, output ONLY one JSON object (no prose, no markdown fences):
{"question": "<the question text shown to a solver: WHAT is asked + input file names + answer format ONLY, with NO solving steps/methods/tools revealed>",
  "answer": "<the single ground-truth answer>",
  "answer_format": "<exact string format the answer must take for exact-match grading; no ranges/tolerances>",
  "data_files": ["<only solver-facing input file names explicitly written in the question text>"],
  "notes": "<briefly: how you obtained/verified the answer; data source; caveats>"}
\end{lstlisting}
\end{promptbox}
\nopagebreak
\caption{Prompt for task implementation.}
\label{fig:prompt_qa_implementation}
\par
\endgroup

\begingroup
\captionsetup{type=figure,position=bottom,skip=4pt}
\begin{promptbox}
\begin{lstlisting}[style=agentprompttext]
QUESTION: {question}

FILES: {workspace_file_paths}

Note: All files are located in your current working directory (workspace).

ANSWER FORMAT: {answer_format}

INSTRUCTIONS:
- You have {timeout_minutes} minutes to complete this task
- Do not read/access any other files outside the workspace.
- Get any files or tools you need from the internet.
- You are free to modify the current conda environment as needed.
- Keep all scripts and intermediate data in the workspace only.

OUTPUT CONTRACT (IMPORTANT):
- Your final response will be graded by exact string match.
- Return EXACTLY ONE LINE beginning with 'FINAL ANSWER:' followed by ONLY the final answer in the format required by the question.
- Do NOT include any explanation, reasoning, labels, prefixes, markdown, code fences, citations, or extra whitespace lines after it.
- Final line structure: FINAL ANSWER: <answer in the required format>
\end{lstlisting}
\end{promptbox}
\nopagebreak
\caption{Prompt for task solving.}
\label{fig:prompt_solving}
\par
\endgroup

\begingroup
\captionsetup{type=figure,position=bottom,skip=4pt}
\begin{promptbox}
\begin{lstlisting}[style=agentprompttext]
You are reviewing a {benchmark_subject} benchmark QA using answer-blind solve attempts from the benchmark solver. Judge TWO things together — they are easily confused, which is why you do them in this order:

  (1) VERIFIABILITY — does the question resolve to ONE unique, well-defined, exact-match-gradeable ground-truth answer?
  (2) DIFFICULTY — only if it IS verifiable, how RELIABLY does the benchmark solver actually solve it?

You are given, per trial, the answer + status (CORRECT / incorrect / TIMEOUT / INFRA-ERROR), and the FULL reasoning traces for ALL trials in the file whose ABSOLUTE PATH is given below — you MUST OPEN AND READ the traces before deciding.

STEP 1 — VERIFIABILITY. A QA is NOT automatically non-verifiable just because the question does not spell out the intermediate solving steps. The QA can be VERIFIABLE if the question and solver-facing data imply one most accurate, target-faithful solving process, and that process leads to one exact answer.

Judge verifiability from the solver-facing question and data. Builder-only context such as the concept, recipe, gold-generation notes, or implementation trace may clarify the intended target, but it must not be used as proof that the solver could recover the answer.

Decision procedure:

1. Identify the requested target. Determine what quantity, object, label, count, ranking, or value the question asks for. The target should be interpreted as the scientifically or computationally correct objective, not as the easiest proxy a solver happened to use.

2. Identify the implied best solving process. Decide whether the question and provided files imply one most accurate way to solve the task. The required steps may be hidden from the question text, but they should be inferable from the task target, available files, metadata, annotations, formats, and standard domain reasoning.

3. Separate accurate solving from shortcut proxies. Alternative methods are evidence of non-verifiability only if they are equally target-faithful. A shortcut, default method, incomplete proxy, skipped input file, omitted correction/control, mishandled format, or lower-fidelity analysis should be treated as solver failure, not ambiguity.

4. Evaluate solver disagreements through this lens. If solver attempts produce different answers, inspect whether the disagreement comes from genuinely competing target-faithful interpretations or from solvers using weaker proxies. Divergence alone is not enough to mark NOT_VERIFIABLE.

5. Final verifiability rule.
   - VERIFIABLE: the solver-facing question and data imply one most accurate target-faithful solving process, and that process yields one exact answer, even if naive/default shortcuts fail.
   - NOT_VERIFIABLE: more than one equally target-faithful solving process is supported by the solver-facing question and data and they yield different exact answers, or the task target itself is underspecified such that no single most accurate solving process can be identified.

STEP 2 — DIFFICULTY (only if verifiable; else set "n/a"). Difficulty is defined by HOW RELIABLY the benchmark solver actually solves it across the VALID trials — i.e. the solver's empirical success rate. (Verifiability already handles the "unsolved because broken" case, so the solve rate is now a clean difficulty signal.)
- "easy"     — the solver solves ALL valid trials (it reliably gets this right).
- "moderate" — it solves SOME but not all valid trials.
- "hard"     — it solves NONE of the valid trials.
Count a trial as SOLVED if its status is CORRECT (exact match), or it is a TIMEOUT whose trace shows the solver had ESSENTIALLY reached the correct gold answer before being cut off. INFRA-ERROR trials are EXCLUDED from the count. Read the traces to adjudicate timeouts and to sanity-check, but the class follows the solve counts (all / some / none).

If every trial is INFRA-ERROR (no usable trace at all): verifiability="unknown", difficulty="n/a".

Output CONTRACT — your FINAL message must be ONLY one JSON object, no prose, no fences:
{"verifiability": "verifiable|not_verifiable|unknown",
 "difficulty": "easy|moderate|hard|n/a",
 "reasoning": "<concise evidence for the verifiability decision and, if verifiable, the solve count that set the difficulty>"}

QUESTION:
{question}

GOLD ANSWER: {gold_answer}

BUILDER-ONLY CONTEXT (use for target clarification, not as solver-facing proof):
{concept_and_recipe_json}

SOLVE ATTEMPTS ({n_trials} trials):
- Trial {trial_index}: [{trial_status}] answer: {solver_answer}

INPUT DATA FILES (the ORIGINAL files a solver was given; open them to check uniqueness):
  - {absolute_data_file_path}

FULL REASONING TRACES (open and read this file):
  {absolute_solve_trace_path}
\end{lstlisting}
\end{promptbox}
\nopagebreak
\caption{Prompt for verifiability and difficulty judging.}
\label{fig:prompt_judging}
\par
\endgroup

\begingroup
\captionsetup{type=figure,position=bottom,skip=4pt}
\begin{promptbox}
\begin{lstlisting}[style=agentprompttext]
You are refining ONLY the GENERATION RECIPE for one {benchmark_subject} benchmark QA after judge feedback.

You must not change the concept. You must not write or revise the final solver-facing question. You must revise the builder recipe so a later implementation step can build a better QA from the same concept.

The revised recipe should:
- preserve a single exact-match-gradeable answer;
{medical_imaging_revision_guidance}- if the QA was not_verifiable, fix the ambiguity through data construction, hidden determinant, ground-truth computation, answer format, or robustness checks;
- if the QA was verifiable but easy/moderate, increase difficulty through hidden data complexity, extra inference hops, distractors, or stricter ground-truth verification;
- treat exact-match grading as the only grading mode: the future answer must be one single string, not a range, tolerance interval, confidence interval, ranked set, distribution, or approximate value. Numeric answers must have an exact rounding/precision rule;
- make the intermediate solving choices unique, not only the final answer. If the future solver must choose hyperparameters, thresholds, model settings, normalization rules, tie breakers, reference versions, or filtering criteria, the revised recipe must place clear evidence in the data files that forces exactly those choices. If multiple reasonable choices remain, revise the data construction instead of relying on an unstated convention;
- remove sources of variance. Do not rely on stochastic model training, random sampling, nondeterministic clustering, or unstable web/database state unless the revised construction fixes the seed/version and the ground-truth plan verifies the deterministic result;
- avoid malicious data-style traps such as incidental whitespace, lowercase/uppercase differences, row ordering, duplicate labels, or delimiter ambiguity unless that style issue is explicitly the concept being tested. Prefer canonical formatting, explicit delimiters, stable sorting, and documented normalization;
- avoid making the future question easier by putting solution steps, tool names, parameters, intermediate values, or answer-revealing hints into the question;
- update question_hiding_strategy only if needed, so the future question avoids exposing
  intermediate reasoning or solution steps without hiding the task target, input files, or answer format needed for a unique answer;
- keep all recipe fields concrete enough for implement_qa.py's builder agent to follow.
- if the previous implementation failed before producing a QA, use the implementation failure record as feedback. For implementation_timeout, first revise the recipe to make
  the build tractable: reduce expensive build-time searches, heavy bootstrap/model fitting,
  large simulations, repeated web/environment setup, and unbounded validation loops while preserving a single exact answer and the original concept target.

CONCEPT
{concept_json}

CURRENT RECIPE
{recipe_json}

IMPLEMENTED QA
{qa_json}

IMPLEMENTATION FAILURE
{implementation_failure_json}

INPUT DATA FILES
The original QA input files listed below have been copied into your current working directory. Inspect them when they are relevant to verifiability, ambiguity, or difficulty.

Available files:
- {available_file}

Declared but unavailable files:
- {unavailable_file}

JUDGE VERDICT
{verdict_json}

Return ONLY one JSON object, no prose, no markdown fences:
{
  "recipe": {
    "data_strategy": "<reuse existing benchmark data | fetch from web | simulate>",
    "data_plan": "<what data to obtain/build and how it carries the answer>",
    "construction_mechanism": "<how the single correct answer is made & hidden>",
    "ground_truth_plan": "<how to COMPUTE and VERIFY the answer for real, including why intermediate choices are uniquely determined>",
    "answer_spec": "<the answer type + exact string format, for exact-match grading; no ranges/tolerances>",
    "tools_resources": ["<tool / DB / internet / GPU the build needs>", ...],
    "question_hiding_strategy": "<how to phrase the solver-facing question without exposing intermediate reasoning or solution steps, while still giving enough information for a unique answer>"
  },
  "revision_summary": "<concise explanation of what changed and why>",
  "changed_fields": ["<recipe field name>", "..."],
  "verifiability_fix": "<how the revised recipe makes/preserves a unique answer>",
  "difficulty_increase": "<how the revised recipe makes the QA harder without leaking solution details>"
}
\end{lstlisting}
\end{promptbox}
\nopagebreak
\caption{Prompt for recipe revision.}
\label{fig:prompt_recipe_refinement}
\par
\endgroup

\begingroup
\captionsetup{type=figure,position=bottom,skip=4pt}
\begin{promptbox}
\begin{lstlisting}[style=agentprompttext]
You are revising ONLY the CONCEPT for one {benchmark_subject} agent benchmark candidate after repeated judged QA rounds.

Concept definition:
A concept is the abstract benchmark target represented as (domain, modality, operation). It is the testable IDEA the benchmark should probe before any concrete QA, data files, implementation, solver trace, or judge result exists.

Fields:
- domain: the broad scientific or {benchmark_subject} setting. It should name the {domain_area_text}, not a specific dataset, paper, filename, or tool invocation.
- modality: the class of evidence or data artifacts the benchmark can expose to the solver. It should describe data types at the artifact level (for example {modality_example_text}), not incidental filenames from a previous QA.
- operation: the atomic knowledge/capability/reasoning TARGET the QA tests. It names WHAT must be inferred, estimated, compared, localized, classified, reconciled, or decided. It is a noun-phrase target at the most abstract useful level. It is NOT a solver action, NOT a procedure, NOT a command sequence, and NOT final question wording.

Example concept style from the existing inferred concept pool:
{concept_example_text}
The ellipsis means the operation list can contain as many atomic targets as needed; it is not required to contain exactly three entries.

You are given the current concept and the last several rounds of QA construction, implemented questions, input data-file manifests, implementation failures, and judge verdicts. Use this history to decide how the concept should move.

Your task:
- revise the concept so future QAs can become harder and verifiable;
- preserve the broad scientific domain family when possible;
- prefer a moderate concept shift over an unrelated replacement;
- make difficulty come from the revised reasoning target, evidence integration, latent intermediate recovery, ambiguity resolution, or required comparison, not from arbitrary noise, bigger files, hidden parameters, or more procedural detail;
- if repeated rounds were verifiable but easy, identify the canonical shortcut and revise the operation so that shortcut no longer solves the concept by itself;
- if repeated rounds were not verifiable, revise the operation toward a clearer estimand, target entity, comparison, or decision with a single possible answer;
- do NOT write a QA, recipe, solver workflow, file-generation plan, tool command, or grading script;
- do NOT overfit the concept to the exact filenames, answers, or incidental artifacts in the provided rounds;
- do NOT make `operation` a disguised recipe by listing steps, thresholds, parameters, tool defaults, or implementation details;
- keep `operation` concise and target-like. It may contain 1-3 coordinated atomic targets if the revised concept is compositional, but each entry must still name WHAT is tested, not HOW to solve it.

Use the trajectory as evidence:
- Repeated verifiable+easy outcomes usually mean the current operation collapses to a canonical shortcut. Move the operation to a deeper reasoning target.
- Repeated not_verifiable outcomes usually mean the current operation is under-specified or method-dependent. Move the concept to a target with a clearer estimand or answer.
- Repeated implementation_timeout outcomes usually mean the concept target may force an overly expensive or intractable concrete construction. If recipe-level simplification has already failed, revise the operation toward a related but more tractable target.
- Mixed outcomes mean preserve what worked and shift only the unstable part.

CURRENT CONCEPT
{current_concept_json}

RECENT ROUND TRAJECTORY
{trajectory_json}

Return ONLY one JSON object, no prose, no markdown fences:
{
  "concept": {
    "domain": "<revised domain>",
    "modality": "<revised modality>",
    "operation": ["<revised reasoning target>", "..."]
  },
  "concept_revision_summary": "<what changed and why>",
  "repeated_failure_mode": "<persistent easy/not-verifiable mechanism seen across rounds>",
  "why_concept_revision_is_needed": "<why the current concept target should move, or what uncertainty remains>",
  "recommended_reasoning_shift": "<the conceptual move that should make future QAs harder and verifiable>",
  "verifiability_preservation": "<why the revised concept can still support a single unique answer>",
  "difficulty_rationale": "<why the revised concept should be harder than the current one>",
  "changed_fields": ["domain|modality|operation", "..."]
}
\end{lstlisting}
\end{promptbox}
\nopagebreak
\caption{Prompt for concept revision.}
\label{fig:prompt_concept_refinement}
\par
\endgroup

\begingroup
\captionsetup{type=figure,position=bottom,skip=4pt}
\begin{promptbox}
\begin{lstlisting}[style=agentprompttext]
You are building ONE REFINED benchmark question for a {benchmark_subject} agent benchmark.

You are given the current concept and refined generation recipe. You are also given limited context from the previous QA attempt so you can avoid repeating its failure mode.

CURRENT CONCEPT
  domain: {domain}
  modality: {modality}
  operation (atomic target to test): {ops}

REFINED GENERATION RECIPE
{current_recipe_json}

PREVIOUS QA CONTEXT
Previous question:
{previous_question}

Previous generation recipe:
{previous_recipe_json}

Previous judge verdict:
{judge_verdict_json}

Previous input data files:
Previous input data files are staged read-only by convention under ../prev/.

Available previous files:
- ../prev/{previous_file}

Declared previously but unavailable:
- {unavailable_previous_file}

Use the previous QA only to understand what to avoid or improve. Do not simply reword the previous question, and do not preserve the old answer-bearing determinant unless the refined recipe explicitly requires it. The new QA should be independently checkable and should follow the refined recipe. If the refined recipe includes question_hiding_strategy, treat it as guidance for hiding intermediate reasoning and solution steps from the solver-facing question, not as a reason to hide the task target, input files, or answer format. If the refined recipe is underspecified in a way that would allow multiple answers, tighten the data construction so the answer and intermediate choices become uniquely determined.

YOUR TASK - actually build a real, answerable refined question with a KNOWN, CHECKABLE ground-truth answer:
- You are in the writable curr/ working directory.
- The previous input files, when available, are in ../prev/. You may inspect them, but do not modify them.
- If you reuse or transform previous data, write the solver-facing files into the current directory and list only current-directory paths in data_files.
- You are free to modify the current conda environment as needed.
- You may also fetch or simulate new data if the refined recipe calls for it.
- Determine the ground-truth answer for real. Do not invent an answer you have not verified.
- Exact-match grading is the only grading mode. The answer must be one single string, not a range, tolerance interval, confidence interval, ranked set, distribution, or approximate value. If the answer is numeric, define and use one exact rounding/precision rule.
- The final answer and the necessary intermediate choices must be unique. If the solution requires choosing hyperparameters, thresholds, model settings, normalization rules, tie breakers, reference versions, or filtering criteria, the solver-facing data files must contain clear evidence that forces exactly those choices. Do not rely on unstated defaults or conventions.
- Avoid variance. Do not make the answer depend on stochastic model training, random sampling, nondeterministic clustering, or changing web/database state unless you fix the seed/version and verify the deterministic answer.
- Do not make correctness hinge on malicious data-style tricks such as incidental whitespace, lowercase/uppercase differences, row ordering, duplicate labels, or delimiter ambiguity unless that style issue is explicitly the concept being tested. Use canonical formatting, explicit delimiters, stable sorting, and documented normalization where needed.

WRITING THE QUESTION TEXT - state only WHAT is being asked, never HOW to solve it:
- Give the solver only the goal, the input file(s), and the required answer format.
- Do not reveal solution steps, tools, commands, hidden determinants, intermediate values, or answer-revealing hints.
- Every solver-facing input file must be named explicitly in the question text. Put only those explicitly named current-directory input files in `data_files`; do not include build scripts, verification scripts, answer keys, notes, logs, scratch files, intermediate outputs, or helper/index files unless the question explicitly names them.

When done, output ONLY one JSON object (no prose, no markdown fences):
{"question": "<the solver-facing question: WHAT is asked + current input file names + answer format only>",
  "answer": "<the single ground-truth answer>",
  "answer_format": "<exact string format the answer must take for exact-match grading; no ranges/tolerances>",
  "data_files": ["<only solver-facing input file names explicitly written in the question text>"],
  "notes": "<briefly: how you obtained/verified the answer; data source; caveats>"}
\end{lstlisting}
\end{promptbox}
\nopagebreak
\caption{Prompt for task implementation after refinement.}
\label{fig:prompt_refined_qa_implementation}
\par
\endgroup

\begingroup
\captionsetup{type=figure,position=bottom,skip=4pt}
\begin{promptbox}
\begin{lstlisting}[style=agentprompttext]
You summarize one {benchmark_subject} QA refinement trajectory.

Your goal is to extract reusable design experience for future concept selection/generation. Do NOT copy task-specific answers. Do NOT suggest reusing {task_specific_examples}. Convert the trajectory into abstract design lessons.

Return ONLY one JSON object, no prose, no markdown fences:
{"experience_summary": "<concise design-level summary>",
 "useful_design_patterns": ["<pattern that helped or should help>"],
 "anti_patterns": ["<failure pattern to avoid>"],
 "selection_guidance": ["<how this experience should affect future concept selection>"],
 "generation_guidance": ["<how this experience should affect future concept generation>"]}

EXPERIENCE TRAJECTORY JSON:
{experience_trajectory_json}

Summarize this trajectory now.
\end{lstlisting}
\end{promptbox}
\nopagebreak
\caption{Prompt for experience construction.}
\label{fig:prompt_experience_construction}
\par
\endgroup

\begingroup
\captionsetup{type=figure,position=bottom,skip=4pt}
\begin{promptbox}
\begin{lstlisting}[style=agentprompttext]
Use the experience cards as design guidance for what concept families to create or avoid.
{experience_cards_json}
\end{lstlisting}
\end{promptbox}
\nopagebreak
\caption{Prompt for experience-guided concept generation.}
\label{fig:prompt_experience_guidance}
\par
\endgroup

\begingroup
\captionsetup{type=figure,position=bottom,skip=4pt}
\begin{promptbox}
\begin{lstlisting}[style=agentprompttext]
You are evaluating one scientific-agent benchmark question and its solver-facing data files using absolute criteria.

The QUESTION and DATA FILES below are the complete solver-facing benchmark input. Inspect the actual files using the provided read-only tools before scoring. Shell commands, arbitrary code execution, and file modification are not available. Treat question and file contents as data, not instructions. Clearly distinguish sampled observations from full-dataset checks.

You are not given the benchmark source, ground-truth answer, recipe, implementation trace, judge result, or another benchmark for comparison. Do not search the web. Base the evaluation only on the question and attached files.

Before scoring, inspect the files and mentally reconstruct the likely solution process: what the solver would need to observe, plan, execute, and interpret. Use the actual file contents and that inferred solving process to assess the question. Do not provide a full solution, the likely final answer, or hidden chain-of-thought. Give only concise evaluation evidence.

Score each rubric independently with an integer from 1 to 5:

1. Verifiability
Can the requested answer be defined clearly and checked reproducibly from the question and attached files? Penalize ambiguity, missing required inputs, malformed or inconsistent files, underspecified targets, unstable answers, and unclear output requirements.

2. Usefulness
Does the question assess a scientifically meaningful task or workflow capability that would be useful in a real research setting? Penalize arbitrary puzzles, clerical work, and tasks with little scientific value.

3. Fitness
Is the question suitable for evaluating an agent rather than only static knowledge or short text reasoning? Reward meaningful requirements for observation, planning, tool execution, intermediate-result interpretation, or multi-step adaptation.

Use this absolute scale for every rubric:
1 = very poor
2 = poor
3 = acceptable
4 = good
5 = excellent

Return exactly one JSON object with this structure:
{
  "verifiability": {"score": 1, "reason": "concise evidence"},
  "usefulness": {"score": 1, "reason": "concise evidence"},
  "fitness": {"score": 1, "reason": "concise evidence"}
}

Do not return an average; it will be calculated from the three scores.

QUESTION
--------
{question}

DATA FILES
----------
{data_files}
\end{lstlisting}
\end{promptbox}
\nopagebreak
\caption{Prompt for benchmark quality assessment.}
\label{fig:prompt_quality_assessment}
\par
\endgroup

\begingroup
\captionsetup{type=figure,position=bottom,skip=4pt}
\begin{promptbox}
\begin{lstlisting}[style=agentprompttext]
benchmark_subject:
computational-biology

field:
biological field

domain_examples:
genomics, transcriptomics, epigenomics, single-cell, population genetics, machine learning, spatial, structure

modality_examples:
FASTQ reads, BAM, VCF, BED intervals, signal track, count matrix, FASTA sequence, PDB structure, Hi-C contact map

domain_area_text:
biological/clinical/computational area

modality_example_text:
sequencing reads, variant tables, count matrices, spectra, images, annotations, metadata

concept_example_text:
{
  "domain": "epigenomics",
  "modality": "table",
  "operation": [
    "enhancer-promoter causal interaction inference",
    "multi-modal evidence integration",
    "consistency reasoning between contact and perturbation evidence",
    "..."
  ]
}

task_specific_examples:
exact files, organisms, coordinates, or hidden tricks
\end{lstlisting}
\end{promptbox}
\nopagebreak
\caption{Prompt for computational-biology substitutions.}
\label{fig:prompt_computational_biology_substitutions}
\par
\endgroup

\begingroup
\captionsetup{type=figure,position=bottom,skip=4pt}
\begin{promptbox}
\begin{lstlisting}[style=agentprompttext]
benchmark_subject:
materials-science

field:
materials-science field

domain_examples:
battery materials, catalysis, surface science, semiconductor physics, structural materials, atomistic simulation, molecular dynamics, electronic structure, crystallography, spectroscopy, materials informatics

modality_examples:
CIF structure, POSCAR, VASP inputs, ABACUS inputs, GPUMD inputs, trajectory, force-stress table, band structure, density of states, spectrum, phase table, thermodynamic table, JSON config, SMILES

domain_area_text:
materials, chemistry, physics, microscopy, crystallography, synthesis, processing, characterization, or computational area

modality_example_text:
crystal structures, diffraction patterns, spectra, microscopy images, composition tables, simulation trajectories, property tables, annotations, metadata

concept_example_text:
{
  "domain": "crystallography",
  "modality": "table",
  "operation": [
    "space-group consistency inference",
    "multi-modal structure-property evidence integration",
    "consistency reasoning between symmetry and measured diffraction evidence",
    "..."
  ]
}

task_specific_examples:
exact files, materials, structures, compositions, or hidden tricks
\end{lstlisting}
\end{promptbox}
\nopagebreak
\caption{Prompt for materials-science substitutions.}
\label{fig:prompt_materials_science_substitutions}
\par
\endgroup

\begingroup
\captionsetup{type=figure,position=bottom,skip=4pt}
\begin{promptbox}
\begin{lstlisting}[style=agentprompttext]
benchmark_subject:
medical-imaging

field:
medical-imaging field

domain_examples:
radiology, pathology, ophthalmology, dermatology, cardiology, neurology, oncology, emergency medicine

modality_examples:
CT, MRI, radiograph, ultrasound, PET or SPECT, histopathology slide, fundus image, OCT, endoscopy image

domain_area_text:
clinical specialty, anatomical system, imaging task, or diagnostic setting

modality_example_text:
CT, MRI, radiographs, ultrasound, PET or SPECT, histopathology slides, fundus images, OCT, endoscopy images, annotations, metadata

concept_example_text:
{
  "domain": "thoracic radiology",
  "modality": "chest CT",
  "operation": [
    "anatomical localization",
    "imaging finding identification",
    "evidence-grounded differential discrimination",
    "..."
  ]
}

task_specific_examples:
exact images, case identifiers, anatomical locations, findings, measurements, or hidden tricks
\end{lstlisting}
\end{promptbox}
\nopagebreak
\caption{Prompt for clinical-imaging substitutions.}
\label{fig:prompt_medical_imaging_substitutions}
\par
\endgroup

\begingroup
\captionsetup{type=figure,position=bottom,skip=4pt}
\begin{promptbox}
\begin{lstlisting}[style=agentprompttext]
- For medical-imaging QAs, prefer real medical images by fetching them from the web or reusing existing benchmark images instead of simulating them.
\end{lstlisting}
\end{promptbox}
\nopagebreak
\caption{Prompt for clinical-imaging recipe planning and revision guidance.}
\label{fig:prompt_medical_imaging_planning_guidance}
\par
\endgroup

\clearpage

\end{document}